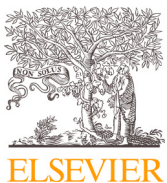
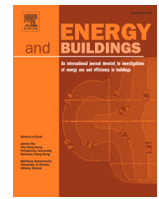

# High-resolution weather-guided surrogate modeling for data-efficient cross-location building energy prediction

Piragash Manmatharasan [a],*, Girma Bitsuamlak [b], Katarina Grolinger [a]

[a] *Department of Electrical and Computer Engineering, Western University, London, N6A 5B9, Ontario, Canada*
[b] *Department of Civil and Environmental Engineering, Western University, London, N6A 5B9, Ontario, Canada*

A B S T R A C T

Building design optimization often depends on physics-based simulation tools such as EnergyPlus, which, although accurate, are computationally expensive and slow. Surrogate models provide a faster alternative, yet most are location-specific, and even weather-informed variants require simulations from many sites to generalize to unseen locations. This limitation arises because existing methods do not fully exploit the short-term weather-driven energy patterns shared across regions, restricting their scalability and reusability. This study introduces a high-resolution (weekly) weather-informed surrogate modeling approach that enhances model reusability across locations. By capturing recurring short-term weather-energy demand patterns common to multiple regions, the proposed method produces a generalized surrogate that performs well beyond the training location. Unlike previous weather-informed approaches, it does not require extensive simulations from multiple sites to achieve strong generalization. Experimental results show that when trained on a single location, the model maintains high predictive accuracy for other sites within the same climate zone, with no noticeable performance loss, and exhibits only minimal degradation when applied across different climate zones. These findings demonstrate the potential of climate-informed generalization for developing scalable and reusable surrogate models, supporting more sustainable and optimized building design practices.

## 1. Introduction

Buildings are among the most significant consumers of energy worldwide and a major source of carbon emissions, contributing to 37% of the global energy-related greenhouse gas output [1]. As such, optimizing the energy performance of buildings plays a critical role in addressing the pressing challenge of climate change. Thoughtful building design can considerably reduce energy consumption through strategies that minimize losses and enhance passive energy gains from natural resources such as solar radiation and ventilation [2]. These design-stage interventions often yield greater long-term sustainability benefits compared to operational-stage strategies such as fault detection, load forecasting, or advanced control systems [3–5].

However, building design optimization remains a computationally intensive task. Assessing the performance of a proposed design typically requires detailed simulations using high-fidelity, physics-based tools such as EnergyPlus [6], DesignBuilder [7], and TRNSYS [8]. These tools accurately model complex interactions among design parameters, environmental conditions, and occupant behavior, delivering robust estimates of building energy performance. Despite their accuracy, the computational cost of these simulations poses a significant challenge, particularly in iterative optimization workflows that demand rapid evaluation of numerous design alternatives.

Surrogate models have emerged as a promising solution to alleviate the computational demands of simulation-based building design optimization [9–12]. These models function as computational proxies, trained to map input design variables to corresponding simulation outputs, thereby enabling rapid performance evaluations that effectively reduce reliance on time-intensive physics-based simulations [13]. By accelerating evaluation, surrogate models shorten iterative optimization cycles. However, their development introduces its own set of challenges. A substantial number of high-fidelity simulations are typically required to generate the training dataset, and additional computational resources are needed to train the model itself. Moreover, surrogate models are typically tailored to a specific building and climate expressed with a weather file, limiting their applicability to new designs or different weather files. As noted by Westermann et al. [14] and Manmatharasan et al. [15], surrogate-assisted frameworks are often developed in a

---

one-off fashion and require retraining for each new building or weather file. This lack of reusability undermines scalability and poses a barrier to broader adoption in real-world design optimization workflows.

Reusing surrogate models presents a promising avenue for improving the efficiency and scalability of building design optimization workflows. When transitioning between optimization tasks, the primary sources of variation are typically the weather file and the building design. In scenarios where the building architecture remains fixed, differences in local climate, captured through weather files, become the dominant factor influencing energy performance. Most existing surrogate models are location-specific and trained to predict energy without using any features related to weather [14–16]. As a result, these models are generally climate-agnostic, and their predictive capabilities do not transfer well across different locations.

While a few prior studies have explored embedding climate awareness into surrogate models-highlighting their potential to improve both reusability and scalability-these approaches rely on simulation data from a large number, or even all, of the required locations to achieve acceptable generalization. This requirement increases computational cost and undermines scalability. Moreover, the use of annual weather data-whether directly as aggregated features or through automated feature engineering-leads to annual-level aggregations that obscure important temporal dynamics [17]. This results in features with limited variability across locations, which restricts the model's ability to learn robust, transferable patterns.

This study addresses the scalability limitations of existing surrogate modeling approaches by introducing a high-resolution, weather-guided surrogate modeling approach designed for cross-location generalization with minimal simulation effort. Unlike conventional surrogates that are location-specific, and unlike transferable models that still require simulation data from hundreds of locations, our approach enables zero-shot prediction in unseen locations using training data from only a few representative sites. This is achieved by learning fine-grained weather-energy relationships through weekly rather than aggregated annual energy prediction. We further analyze the impact of intra and inter climate zone representation in the training data on model transferability. Additionally, we evaluate three time-series learning strategies for encoding the weather input: Temporal Convolutional Networks (TCNs), Transformer-based encoders, and convolutional autoencoders. Through this comparative evaluation, the study identifies data and modeling techniques best suited for enabling scalable and transferable surrogate modeling.

This paper is organized as follows. Section 2 reviews related work. Section 3 presents the methodology. Section 4 reports the experimental results. Section 5 provides a discussion of key findings, and Section 6 concludes the paper.

## 2. Related work

Surrogate models have become a foundational tool in simulation-based building design, enabling the efficient approximation of performance outcomes. A substantial body of research has focused on enhancing model accuracy. To gain acceptable surrogate accuracy, studies explored techniques such as Feed Forward Neural Networks (FFNNs) [18,19], Random Forests (RFs) [20,21], and Gradient-Boosted Machines (GBMs) [22,23]. To further enhance predictive performance, many studies incorporate hyperparameter tuning, using various optimization algorithms including Genetic Algorithms (GAs), Grey Wolf Optimization (GWO), and Particle Swarm Optimization (PSO), to systematically calibrate model structure [24–26]. Ensemble methods have also been explored to increase the robustness of the predictions as they reduce the variances [27–29]. To boost long-term reliability, several studies have adopted climate change-sensitive strategies by incorporating weather inputs derived for climate scenarios in the future years using morphing techniques or General Circulation Models (GCMs)-thereby enhancing the correctness of surrogate predictions under evolving environmental conditions [30,31].

Complementing accuracy improvements, a parallel line of research has focused on improving the efficiency of surrogate modeling pipelines. Adaptive sampling strategies have been used to reduce the number of simulations by selecting design variable combinations that maximize information gain, which reduces the overall computational cost [32,33]. Feature selection methods have also been widely used to reduce input dimensionality, thus improving both model interpretability and training speed [34,35]. At a workflow level, researchers have proposed integrated simulation-surrogate systems that streamline the entire process-from design variable generation to simulation execution and model training-within automated pipelines [36,37]. In parallel, increasing attention has been directed toward model explainability, as complex machine learning surrogates become harder to interpret. Tools such as SHAP values, LIME, and Sobol-based sensitivity analysis have been applied to quantify variable importance and provide insight into surrogate model behavior, enhancing transparency in decision-making contexts [38,39].

Despite these advancements, most surrogate modeling pipelines are developed in a single-scenario fashion, meaning they are trained and applied for a specific building and weather file. As highlighted in the systematic review by Manmatharasan et al. [15], as well as the review by Westermann et al. [14], the lack of generalizability across multiple weather files or designs limits broader application. Scalable surrogate models that can generalize across different designs or weather conditions can lead to broader application of sustainable optimization.

Early studies aiming to develop surrogate models for predicting building performance across varied geographic locations often constructed separate regression models for each location to predict energy consumption independently [40,41]. However, maintaining and training location-specific models is inefficient and not scalable. To address this, later studies began incorporating weather features into a unified model, enabling the use of a single surrogate across multiple locations. For example, Rackes et al. [42] used statistical summaries of weather data to build a surrogate model applicable to 426 Brazilian locations. García et al. [43] introduced a categorical location variable into a hybrid neuro-genetic surrogate framework, achieving generalization across three distinct locations in Mexico. Similarly, Zheng et al. [44] engineered weather features such as annual solar irradiance and temperature averages, allowing their FFNN-based model to generalize across five Chinese locations without retraining. However, these models were trained using data from all intended deployment locations and were not evaluated on any unseen climates. As a result, while they performed well within the training regions, their ability to generalize to entirely new locations remains untested.

The approach by Westermann et al. [17] demonstrated that generalization to unseen locations is possible. They proposed a deep learning framework combining TCNs with FFNNs, trained on hourly weather sequences from 569 Canadian locations. Their model achieved a Mean Absolute Percentage Error (MAPE) below 3% for unseen locations from Canada, showcasing strong cross-location generalization. However, this came at a significant computational cost, requiring extensive simulation data from all 569 locations.

Another key limitation across all the reviewed studies lies in their reliance on low-resolution weather representations, either by using pre-aggregated annual weather metrics or by applying automated feature extraction techniques (e.g., TCNs) over full-year sequences. In both cases, the resulting features often capture only broad, long-term trends while failing to retain shorter-term temporal dynamics such as weekly variability, seasonal shifts, or extreme events. Westermann et al. [17] observed that the features automatically extracted by their TCN model were highly correlated with annual metrics. As a result, these features exhibit limited variation across different locations, which hinder the surrogate models' ability to learn transferrable granular, climate-sensitive patterns. Addressing this gap requires the use of higher-resolution weather inputs (e.g., weekly or daily segments) to enable more robust and transferable surrogate models across diverse climatic contexts.

Such remains a significant opportunity to develop climate-informed surrogate models that generalize to new weather profiles using simulation data from only one or two representative locations. Such models could be reused across different locations with little or no retraining, enabling low-cost, scalable prediction workflows that support iterative design evaluation. This is especially practical in cases where the building geometry remains fixed and only a constrained set of design variables-such as insulation, glazing, or system efficiency-is optimized. Our approach fills this gap by leveraging limited-location data to train surrogates that retain transferability across climates, offering a practical solution where existing methods fall short.

## 3. Methodology

This paper proposes a climate-aware surrogate modeling framework that enables generalization across different locations using simulation data from only one or a few representative locations. This reduces dependence on a large multi-location dataset while improving predictive accuracy across locations. To achieve this, our model predicts weekly energy consumption using weekly weather data rather than relying on features derived from full-year weather inputs. Three different time-series encoding architectures are integrated into this surrogate model, each aiming to extract representative weather embeddings to examine their effectiveness in various use cases.

The following subsections present the methodology in four parts: (1) motivation for weekly resolution energy prediction (2) the generation of simulation-based training data, (3) the mathematical formulation of the surrogate modeling approach, (4) the alternative strategies used to encode time-series weather data into embeddings, and (5) the evaluation metrics used to assess model performance.

### 3.1. Motivation for weekly-resolution surrogate modeling

Weekly resolution provides a more fine-grained and diverse representation of weather variability compared to annual resolution, which is critical for enabling cross-location generalization. Fig. 1 illustrates weekly (smaller, lighter circles) and annual (larger, darker circles) averages of dry-bulb temperature and relative humidity across ten Canadian locations. Annual averages form tightly clustered, low-variance points that are distinct across locations, whereas weekly data exhibit a more dispersed and overlapping distribution. The variance is substantially higher for weekly averages (241.20) compared to annual averages (48.49), indicating a richer and more informative dataset. This increased variability is especially beneficial when training deep learning models with limited weather files, as it enables learning of transferable weather-energy relationships rather than overfitting to location-specific climate summaries. The overlapping nature of weekly patterns further suggests the presence of shared climatic behaviors, increasing the likelihood of discovering generalizable representations.

To further support this insight, we conducted a similarity analysis between two climatically distinct cities, Toronto and Calgary. The cosine similarity between their full annual weather time series was 0.52. However, when weather data were compared at the weekly level and each week was matched to its most similar counterpart in the other city, the average similarity increased to 0.64. These findings offer both intuitive and quantitative justification for adopting weekly-resolution surrogate modeling.

### 3.2. Training data generation through simulation

Simulations play a central role in this study and are performed by systematically varying key building design parameters to generate diverse performance scenarios. To efficiently sample the design space, we employ Latin Hypercube Sampling (LHS)-a stratified statistical technique that ensures uniform coverage across the multidimensional input space by dividing each variable's range into equal intervals. LHS is widely recognized in the literature as a robust and effective method for building performance simulation and has been frequently highlighted in review studies as one of the most commonly adopted strategies for surrogate modeling and design optimization [15]. The parameters varied in this study include thermal, geometric, and orientation-related design features. The parameters and ranges are as follows:

- **Wall insulation thickness:** 0.02-0.10 m
- **Roof insulation thickness:** 0.02-0.10 m
- **Window U-factor:** 1.2-2.0 W/m$^2$K
- **Window SHGC:** 0.3-0.7
- **Visible transmittance:** 0.5-1.0
- **Wall thickness:** 0.1-0.5 m
- **Roof thickness:** 0.1-0.5 m
- **North axis orientation:** 0-360°
- **Wall absorptance:** 0.5-0.9
- **Roof absorptance:** 0.5-0.9
- **Internal equipment gain:** 5-15 W/m$^2$
- **Window scale factor:** 0.5-1.0
- **Heating setpoint:** 18-22°C
- **Cooling setpoint:** 24-28°C

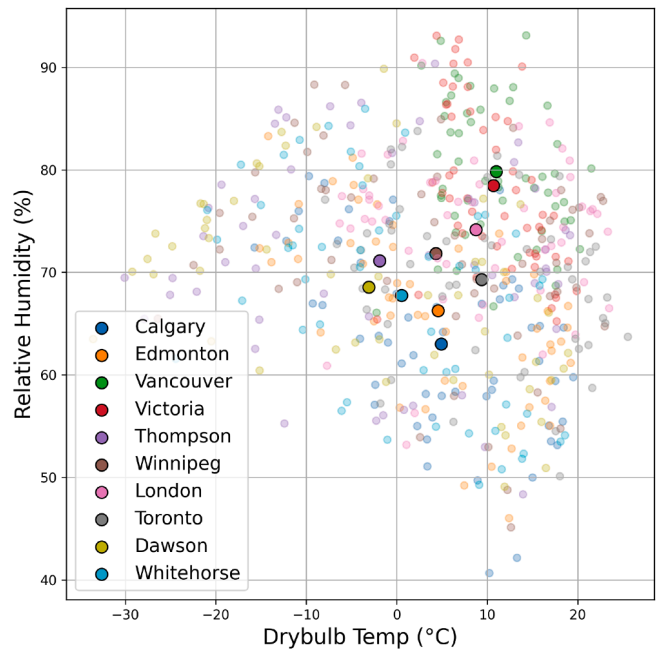

**Fig. 1.** 2D feature space showing drybulb temperature and relative humidity for both weekly (smaller, lighter circles) and annual (larger, darker circles) averages across ten locations.

For each sampled combination of design variables, EnergyPlus simulations were conducted to obtain weekly energy consumption values, yielding 52 predictions (one per week) per design across a full year. This structure enables the model to learn how a given building design responds to diverse weekly weather patterns by leveraging the variability present within a full year of weather data from a single geographic location. By training on weekly segments rather than aggregated annual trends, the model captures short-term temporal dynamics - such as heatwaves, cold spells, and seasonal transitions - that are critical for performance prediction. As a result, it can infer how the same design might behave under different locations, making it possible to generalize without requiring location-specific simulations during training.

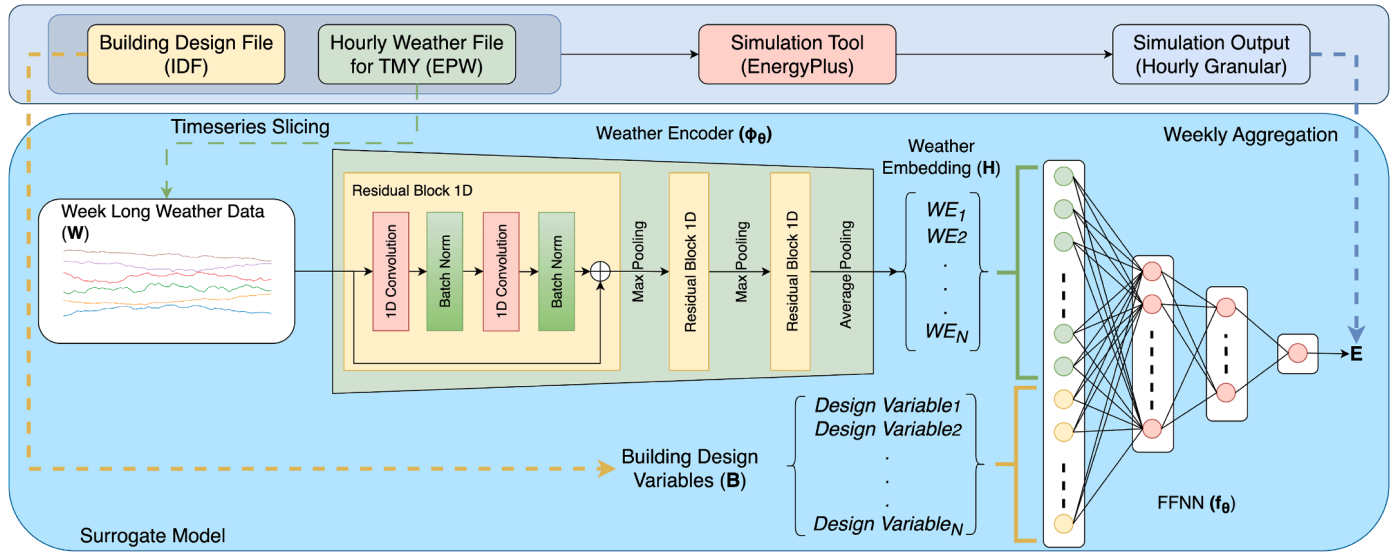

**Fig. 2.** Temporal encoder architecture using a TCN (Approach 1). While both TCN and transformer encoders are used in our study, only the TCN variant is shown; the Transformer serves as a direct replacement within the same structure.

*3.3. Model formulation*

Let $\mathbf{W} \in \mathbb{R}^{T \times d_w}$ represent the time-series weather data for a given week, where $T$ is the number of timesteps and $d_w$ is the number of weather-related features. Let $\mathbf{B} \in \mathbb{R}^{d_b}$ denote the vector of building design variables. The proposed approach estimates the weekly energy consumption $\hat{E} \in \mathbb{R}$ using a weather encoder ($\phi_\theta(\cdot)$) and a FFNN ($f_\theta$):

$$\mathbf{H} = \phi_\theta(\mathbf{W}) \quad (1)$$

$$\hat{E} = f_\theta(\mathbf{H}, \mathbf{B}) \quad (2)$$

Weather encoder $\phi_\theta(\cdot)$ transforms the weather input sequence $\mathbf{W}$ into a compact embedding $\mathbf{H}$, capturing salient weather patterns. This embedding is concatenated with the building design vector $\mathbf{B}$ and passed to the FFNN predictor ($f_\theta$) for final energy consumption prediction $\mathbf{E}$. The following subsections describe two alternative strategies for the weather encoder $\phi_\theta(\cdot)$. Approach 1 employs a joint training strategy, where the encoder and predictor are optimized together to directly align the weather representation with the prediction objective. In contrast, Approach 2 decouples the training process: a weather encoder is first trained independently to produce embeddings for each weekly weather sequence, which are then combined with the building design variables in a separately trained predictor to estimate energy consumption.

*3.4. Approach 1: Joint weather encoding using TCNs and transformers*

Before training, data preprocessing is applied to ensure consistency and support transferability across different locations. Building design variables are normalized using min-max scaling based on their predefined value ranges. Weather variables were normalized using a min-max scaler fitted on the training location weather file, without applying clipping during transformation. This allows weather values from unseen locations to naturally exceed the training range when necessary, preserving relative magnitude differences under more extreme or unfamiliar climatic conditions. While an earlier version of the methodology considered using unscaled weather inputs, the adopted min-max scaling without clipping provides a better balance between numerical stability and physically consistent extrapolation across diverse climates. Each training sample represents one week, consisting of 168 hourly timesteps.

As illustrated in Fig. 2, this proposed approach uses a hybrid architecture where the time-series weather encoder and FFNN are trained jointly to predict weekly energy consumption $\mathbf{E}$. The encoder learns to extract task-specific weather representations $\mathbf{H}$ from the input sequence $\mathbf{W}$, which are then concatenated with the building design variables $\mathbf{B}$ to form the FFNN model input. We examine two deep learning architectures for the weather encoder component: TCNs and Transformers. TCNs have demonstrated strong performance across various time-series tasks due to their ability to model long-range dependencies, while Transformers, equipped with self-attention mechanisms, offer advantages by dynamically weighting temporal patterns, making them particularly suitable for capturing complex weather dynamics.

**TCN based Encoder:** For the TCN-based architecture, Fig. 2, the weather encoder consists of multiple residual 1D convolutional blocks, each containing two convolution layers with batch normalization. These blocks are designed to capture temporal dependencies in the weekly weather sequence while preserving gradient flow through residual connections. Temporal downsampling is performed using max-pooling layers after each block, allowing the network to progressively reduce the sequence length and extract increasingly abstract temporal features. A final global average pooling layer compresses the output into a fixed-length embedding that summarizes the entire weather sequence.

**Transformer based Encoder:** For the Transformer-based architecture, the weather encoder, made up of transformer blocks, replaces the 1D residual convolution blocks. In this architecture, the weekly weather input $\mathbf{W}$ is first passed through a linear embedding layer that maps each timestep to a higher-dimensional feature space. To preserve the sequential information, sinusoidal positional encoding is added to the embedded sequence. The combined input is then passed through a stack of encoder blocks, each composed of multi-head self-attention mechanisms and feedforward sublayers. These blocks allow the model to learn long-range dependencies and dynamically attend to different parts of the weather sequence. The final sequence output is averaged across the time dimension to produce a fixed-length embedding that captures the overall climatic characteristics of the week.

The fixed-length embedding output from the weather encoder ($\mathbf{H}$), TCN or Transformer-based, is concatenated with the building design vector ($\mathbf{B}$) to form a composite representation that jointly encodes climatic and design information. This combined input is passed through an FFNN responsible for predicting weekly energy consumption. The FFNN consists of four sequential linear layers with progressively decreasing dimensionality with a final scalar output layer. Each linear transformation is followed by a ReLU activation function and dropout regularization to mitigate overfitting.

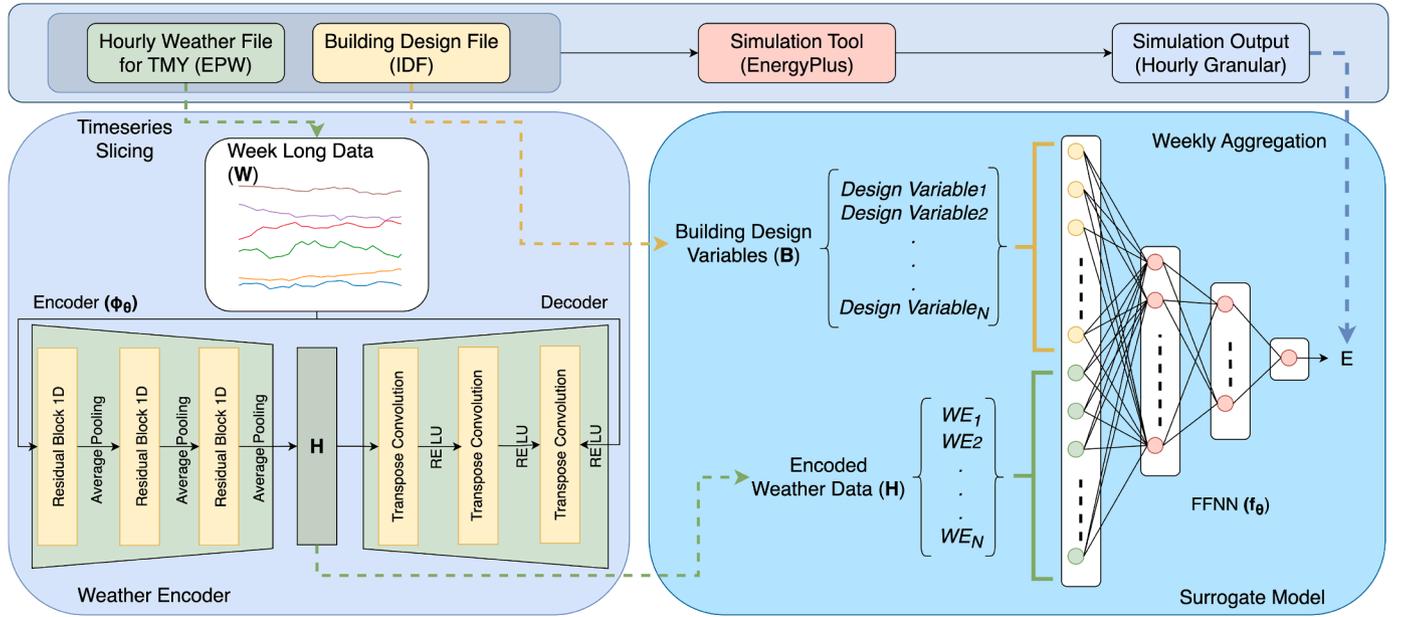

**Fig. 3.** Weather embedding generation using a pre-trained autoencoder (Approach 2). The embedding extractor (autoencoder) is trained separately; this decoupled structure enables the use of multi-location weather data without simulation and reduces training time for downstream prediction models.

The full architecture is jointly trained end-to-end by minimizing the Mean Squared Error (MSE) loss:

$$\mathcal{L}_{\text{pred}} = \frac{1}{N} \sum_{i=1}^{N} \left( E_i - \hat{E}_i \right)^2 \qquad (3)$$

where $E_i$ is the true energy consumption from simulation for the respective week, $N$ is the batch size, and $\hat{E}_i$ is the model prediction. The error in energy prediction is backpropagated from the output layer through the FFNN and further through the weather encoder. This end-to-end training process allows the model to jointly optimize both the encoder and the FFNN by updating their weights based on the prediction loss. Key architectural hyperparameters are optimized to ensure robust performance across diverse climates in both the TCN- and Transformer-based architectures.

### 3.5. Approach 2: Pre-trained weather autoencoder for modular embedding

Data preprocessing is performed in the same manner as in Approach 1, ensuring consistency across both modeling strategies. As shown in Fig. 3, this approach decouples weather feature learning from the prediction task. By decoupling the encoder from the surrogate prediction model, we enable the encoder to be trained once across multiple weather files, significantly reducing simulation overhead and forward pass time for downstream models. This design also enables flexible learning strategies-such as incremental training and multi-location adaptation-in a time-efficient manner. An autoencoder setup is well-suited for this purpose, as it can be trained in an unsupervised manner. Once trained, the encoder can generate relatively accurate embeddings for any location, as it is exposed to a broader range of weather patterns during training-unlike Approach 1, which is limited to weather data only from locations used in simulation-based training.

In this approach, a convolutional autoencoder is first trained to compress weekly weather sequences $\mathbf{W}$ into compact embeddings $\mathbf{H} = \phi_\theta(\mathbf{W})$. The encoder consists of stacked 1D residual convolutional blocks, each comprising two convolution layers followed by batch normalization. Max-pooling layers are used after each block to progressively reduce the temporal dimension. The decoder mirrors this structure using transposed 1D convolutional layers for upsampling, reconstructing the original input sequence from the latent embedding. The model is trained to minimize the reconstruction loss:

$$\mathcal{L}_{\text{rec}} = \frac{1}{N} \sum_{i=1}^{N} \left\| \mathbf{W}_i - \hat{\mathbf{W}}_i \right\|^2 \qquad (4)$$

where $\mathbf{W}_i$ is the ground truth weather sequence for the $i^{\text{th}}$ training sample, $\hat{\mathbf{W}}_i$ is the reconstructed output from the autoencoder, and $N$ is the total number of training samples. After training, the encoder $\phi_\theta$ from the autoencoder is frozen and used to generate embeddings $\mathbf{H}$ for any given weekly weather sequence $\mathbf{W}$.

These weather embeddings ($\mathbf{H}$) are concatenated with building-specific input variables $\mathbf{B}$ and passed to a FFNN $f_\theta$ to predict the weekly energy consumption $\hat{E}$. The prediction function is expressed as:

$$\hat{E} = f_\theta(\mathbf{H}, \mathbf{B}) \qquad (5)$$

The first layer maps the combined embedding-building feature vector into a hidden dimension. The remaining layers consist of a fully connected layer with progressively decreasing dimensions, followed by ReLU activations and dropout for nonlinearity and regularization. A final linear layer outputs a single scalar value corresponding to the predicted weekly energy use.

During the training of the prediction model, only the weights of $f_\theta$ are optimized using the prediction loss $\mathcal{L}_{\text{pred}}$ defined in Eq. (3). This modular training strategy offers substantial computational benefits: the encoder $\phi_\theta$ remains fixed and does not require retraining when applied to new locations, allowing the learned weather embeddings to be reused across different tasks or regions. This separation of feature extraction and prediction enhances both the scalability and efficiency of the surrogate modeling framework.

### 3.6. Evaluation metrics

To evaluate predictive accuracy, we use two widely accepted error metrics: Root Mean Squared Error (RMSE) and Symmetric Mean Absolute Percentage Error (SMAPE). RMSE provides a measure of absolute error sensitive to large deviations, while SMAPE quantifies relative error and is scale-independent, making it suitable for comparing performance across different magnitude ranges. They are defined as follows:

$$\text{RMSE} = \sqrt{\frac{1}{n}\sum_{i=1}^{n}(E_i - \hat{E}_i)^2} \tag{6}$$

$$\text{SMAPE} = \frac{1}{n}\sum_{i=1}^{n}\frac{|E_i - \hat{E}_i|}{\frac{|E_i|+|\hat{E}_i|}{2}} \times 100 \tag{7}$$

where $E_i$ is the actual value, $\hat{E}_i$ is the predicted value, and $n$ is the number of data points.

The Pearson correlation coefficient (r) is specifically used to assess how well the models capture the temporal pattern of energy consumption across the 52-week period for a given design configuration. The Pearson correlation coefficient is defined as:

$$r = \frac{\sum_{i=1}^{n}(E_i - \bar{E})(\hat{E}i - \bar{\hat{E}})}{\sqrt{\sum i = 1^n(E_i - \bar{E})^2}\sqrt{\sum_{i=1}^{n}(\hat{E}_i - \bar{\hat{E}})^2}} \tag{8}$$

Where $E_i$ is the actual value, $\hat{E}_i$ is the predicted value, $\bar{E}$ is the mean of the actual values, $\bar{\hat{E}}$ is the mean of the predicted values, and n is the number of data points. A Pearson coefficient close to 1 indicates a strong positive correlation, suggesting that the model effectively captures the variation patterns in energy consumption over the weeks, even if the magnitude of predictions is not perfectly aligned.

Although American Society of Heating, Refrigerating and Air-Conditioning Engineers (ASHRAE) guideline provides widely used calibration benchmarks in terms of Coefficient of Variation of the RMSE (CVRMSE) for monthly (15%) and hourly (30%) energy model validation [45,46], there is no direct benchmark defined for RMSE or SMAPE, particularly in the context of surrogate modeling for design optimization. Accordingly, both metrics are interpreted qualitatively in this study, where values closer to zero indicate higher predictive fidelity and greater suitability for optimization purposes. Maintaining low relative error is essential, as insufficient surrogate accuracy can distort the relative ranking of candidate designs, increasing the risk of convergence toward suboptimal or misleading optimization outcomes.

To support this qualitative interpretation, heatmap visualizations are employed, in which lower SMAPE values are mapped to favorable (green) regions and higher errors to unfavorable (red) regions. This representation enables rapid identification of reliable surrogate configurations during optimization. Furthermore, weekly SMAPE exhibits a broader error distribution, whereas annual aggregation of weekly predictions results in consistently lower SMAPE values across most applicable regions, as expected due to temporal smoothing effects.

## 4. Evaluation

This section presents the results of the evaluation, which assess the generalization ability of the proposed climate-informed surrogate models across diverse geographic locations. We begin by describing the case study building and simulation setup, followed by an explanation of the cross-location evaluation strategy. Results are then organized into two main parts: models trained using data from a single location and those trained on paired-location combinations to enhance diversity. Then, the impact of strategically selecting training locations on model generalization across different climate zones will be examined. Finally, a comparison is conducted between the performance of our weekly prediction strategy and traditional annual-level surrogate modeling approaches. Together, these results provide valuable insights into how different weather embedding strategies, training data diversity, and weekly prediction formulations influence the scalability and generalization of surrogate models.

### 4.1. Case study building and simulation

This study uses the **medium office prototype building** from EnergyPlus (Fig. 4) as the baseline building model for energy performance simulations. This prototype reflects typical commercial and institutional buildings in terms of size, layout, and occupancy patterns, making it a practical choice for evaluating energy-efficient design strategies under varied climatic conditions.

To enable large-scale experimentation, the simulation pipeline was automated using a Python-based workflow. This pipeline generated design variants for each location using LHS, modified the corresponding EnergyPlus input files, and executed simulations programmatically. For each location, 50 design variants of the medium office prototype building were simulated, and each simulation produced 52 weekly energy consumption values, resulting in a 52×50 dataset per location.

This study is specifically focused on investigating the scalability of surrogate models across climatic conditions, using a fixed building geometry and occupancy type. While the medium office prototype provides a representative testbed for commercial buildings, the current approach does not assess generalizability across different building typologies or design archetypes. Therefore, the results reflect the weather-related scalability of the surrogate modeling approaches rather than their ability to generalize across diverse building forms. Future research should extend this work to include residential and heterogeneous buildings to evaluate design-level transferability.

### 4.2. Cross-location evaluation strategy

This subsection outlines the evaluation strategy used to assess the generalizability of the proposed surrogate modeling approaches within a climate zone and across zones. The results presented in the following three subsections follow this structure.

Two types of cross-location experiments are conducted in this study. In the first setup, a model is trained using data from a **single location** and evaluated across the remaining locations to assess its ability to generalize beyond the source location. In the second setup, a model is trained using data from **two distinct locations** and then evaluated across the remaining locations. This pairing helps examine how combining diverse climatic signals during training improves generalization.

To ensure sufficient climatic and geographic diversity, we selected ten Canadian cities grouped across five broad climate zones. These zones were defined based on distinct weather patterns, temperature extremes, and seasonal variability, similar to the ASHRAE climate classification. The zoning captures a progression from mild coastal conditions to extreme subarctic climates, providing a representative spread of building-relevant weather conditions.

Within each zone, two cities were selected based on three criteria: (i) population and urban relevance , (ii) availability of high-quality weather files, and (iii) spatial representativeness within the zone (to avoid localized overfitting and capture intra-zone variability), This selection ensures that each zone is sufficiently represented while allowing for meaningful intra-zone and cross-zone transferability analysis.

- **Zone 1 (Z1):**
  - **Vancouver (Van):** Coastal city with mild temperatures, high humidity, and frequent precipitation.
  - **Victoria (Vic):** Island city with a temperate marine climate, marked by moderate rainfall and minimal snowfall.
- **Zone 2 (Z2):**
  - **Toronto (Tor):** Urban center with a humid continental climate, characterized by hot summers and cold winters.
  - **London (Lon):** Inland city with cold winters and warm summers, representing a typical Southern Ontario climate.
- **Zone 3 (Z3):**
  - **Calgary (Cal):** Prairie city with cold, dry winters and rapid temperature fluctuations due to Chinook winds.
  - **Edmonton (Edm):** Northern prairie city with longer winters and a higher frequency of sub-zero temperatures.

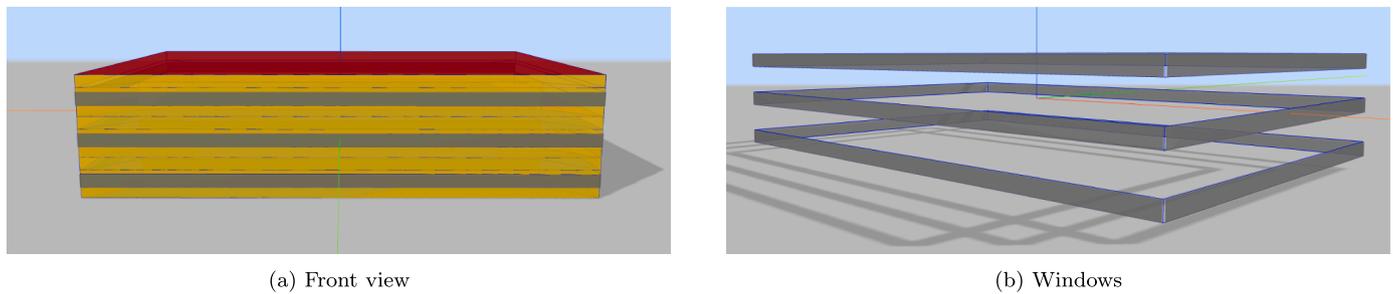

(a) Front view  (b) Windows

**Fig. 4.** Reference building selected from EnergyPlus examples. The geometry is representative of many small-to-medium commercial buildings.

- **Zone 4 (Z4):**
  - **Winnipeg (Win):** Flat prairie city with extreme seasonal temperature variation and dry, frigid winters.
  - **Thompson (Tho):** Northern Manitoba city with prolonged winter periods and subarctic tendencies.
- **Zone 5 (Z5):**
  - **Dawson (Daw):** Remote northern city with very cold winters, low solar exposure, and limited precipitation.
  - **Whitehorse (Whi):** Territorial capital with a cold, dry climate and long-duration snow cover.

This structured selection enables a controlled evaluation of how well surrogate models trained in one or two climates can generalize to others, particularly as climate severity increases from Zone 1 to Zone 5. For any case where the training and test locations are the same, a different Typical Meteorological Year (TMY) weather file from the same location is used for training and testing. Similarly, in each training instance different TMY weather file is used for validation to prevent data leakage.

### 4.3. Single-location training evaluation

This section presents the results of training surrogate models using simulation data from a single location and evaluating their generalization to other locations, both within the same climate zone (intra-zone) and across different climate zones (inter-zone). All three variants-TCN-based, Transformer-based, and Autoencoder-based-were assessed.

The architectural components and tuning ranges were selected to balance the model's capacity to be expressive enough to capture short-term weather variability and longer seasonal dynamics, yet constrained to avoid overfitting to location-specific characteristics. Accordingly, convolutional filter counts, embedding dimensions, and network depths were explored within ranges commonly used in time-series forecasting, providing sufficient representational power without excessive parameterization. Dropout rates and learning rates were tuned conservatively to support stable optimization and promote generalization, particularly when training data are limited to one or two locations.

For the TCN-based models, hyperparameter tuning focused on five key components: the number of filters in the first convolutional block (with subsequent blocks doubling in size), the embedding dimension of the weather encoder, the size of the first hidden layer in the FFNN, the dropout rate, and the learning rate for the Adam optimizer. The best-performing hyperparameter values when each location serves as the training location are listed in Table 1. For the Transformer-based models, tuning included the embedding dimension, number of attention heads, the size of the internal feedforward network within each Transformer block, the FFNN hidden layer size, dropout rate, and learning rate-optimized as detailed in Table 2. All hyperparameter optimization was conducted exclusively using the training location associated with each experiment. Each row in the table therefore corresponds to a distinct training instance in which a single location served as the source of simulation data, and hyperparameters were selected based on validation using an independent weather file from the same location. Once trained and tuned, the resulting model was applied directly to the remaining locations without any further tuning or adaptation. This evaluation protocol reflects a realistic deployment scenario and ensures that performance on unseen locations represents true cross-location generalization rather than location-specific optimization.

For the autoencoder-based approach, tuning is conducted in two stages. First, the weather encoder is tuned using weather data from multiple locations. As this encoder is trained once across the full dataset, a single set of hyperparameters is selected for it. In the second stage, a FFNN prediction layer is tuned independently for each location using the learned embeddings and simulation data. The autoencoder's hyperparameters include the embedding dimension, the number of filters in the initial convolutional block (with subsequent blocks doubling in size), and the learning rate. The selected values are: an embedding dimension of 56, 128 filters in the first convolutional block, and a learning rate of 0.00153. For the FFNN prediction model, the number of hidden layers, the size of the first hidden layer, the dropout rate, and the learning rate are tuned separately for each location. The selected hyperparameters for each FFNN are summarized in Table 3.

**Table 1**
Each row represents a distinct training instance where the TCN-based model was tuned using data from that specific training location. Each of the trained instances is then used to evaluate its performance on other locations, to assess transferability.

| Loc | CNN Filters | Embed Dim | FFNN Size | Dropout | Learning Rate |
|---|---|---|---|---|---|
| Van | 128 | 80 | 256 | 0.178 | 0.00099 |
| Vic | 16 | 64 | 160 | 0.277 | 0.00043 |
| Tor | 32 | 48 | 64 | 0.365 | 0.00080 |
| Lon | 112 | 80 | 64 | 0.142 | 0.00929 |
| Edm | 64 | 16 | 192 | 0.124 | 0.00226 |
| Cal | 16 | 48 | 64 | 0.159 | 0.00508 |
| Win | 48 | 96 | 224 | 0.270 | 0.00386 |
| Tho | 48 | 16 | 96 | 0.355 | 0.00379 |
| Daw | 16 | 64 | 128 | 0.157 | 0.00431 |
| Whi | 48 | 64 | 96 | 0.168 | 0.00090 |

**Table 2**
Each row represents a distinct training instance where the transformer-based model was tuned using data from that specific training location. Each of the trained instances is then used to evaluate its performance on other locations, to assess transferability.

| Loc | Embed Dim | Heads | T-FFNN Size | FFNN Size | Dropout | Learning Rate |
|---|---|---|---|---|---|---|
| Van | 112 | 14 | 87 | 207 | 0.159 | 0.00049 |
| Vic | 16 | 2 | 75 | 154 | 0.204 | 0.00062 |
| Lon | 112 | 14 | 46 | 256 | 0.120 | 0.00068 |
| Tor | 16 | 2 | 46 | 193 | 0.169 | 0.00170 |
| Edm | 88 | 11 | 50 | 192 | 0.141 | 0.00040 |
| Cal | 8 | 1 | 9 | 51 | 0.116 | 0.00060 |
| Win | 16 | 2 | 34 | 121 | 0.179 | 0.00092 |
| Tho | 8 | 1 | 115 | 166 | 0.273 | 0.00103 |
| Daw | 8 | 1 | 100 | 146 | 0.199 | 0.00204 |
| Whi | 32 | 4 | 68 | 32 | 0.150 | 0.00177 |

**Table 3**

Each row represents a distinct training instance where the Autoencoder-based model was tuned using data from that specific training location. Each of the trained instances is then used to evaluate its performance on other locations, to assess transferability.

| Loc | Hidden Dim | Num Layers | Dropout | Learning Rate |
| --- | --- | --- | --- | --- |
| Van | 457 | 3 | 0.289 | 0.00078 |
| Vic | 510 | 4 | 0.235 | 0.00908 |
| Tor | 252 | 5 | 0.344 | 0.00978 |
| Lon | 298 | 5 | 0.448 | 0.00558 |
| Edm | 206 | 5 | 0.430 | 0.00615 |
| Cal | 186 | 5 | 0.198 | 0.00528 |
| Win | 255 | 4 | 0.330 | 0.00991 |
| Tho | 326 | 4 | 0.233 | 0.00891 |
| Daw | 369 | 5 | 0.454 | 0.00978 |
| Whi | 457 | 3 | 0.413 | 0.00556 |

Tables 4 and 5 present the SMAPE and MSE results of the single-location training experiments. In each case, a model trained on one source location is evaluated on three tiers of targets: (i) a different TMY file from the same location, (ii) another location within the same climate zone, and (iii) locations from other climate zones. For each training and testing location pair, the lowest SMAPE score among the three modelling techniques is highlighted. For example, when Vancouver is used as both the source and target location, the Transformer model achieves a SMAPE of 5.48, which is lower than that of the TCN (6.76) and Autoencoder (5.72), and is therefore highlighted in the table. Overall, the Transformer model consistently outperforms both the TCN and Autoencoder-based approaches, likely due to its superior capacity to capture long-range temporal dependencies in time-series weather data.

While these tables provide an overall comparison of accuracy across locations, they make it difficult to observe finer details and do not fully reveal the underlying structure of transferability patterns. To better visualize and interpret these trends, heatmaps are presented in Figs. 5–7, showing the weekly SMAPE heatmaps for all three surrogate modeling approaches. Each heatmap reveals clear evidence of transferability, particularly within the same climate zone. This is expected, as locations within a common climate zone tend to share similar weather patterns and dominant climatic features. As a result, models trained on one location often generalize well to another within the same zone. This pattern is especially visible along the heatmap diagonals, where models are evaluated on different TMY years from the same location or on another location within the same zone-both cases showing lower SMAPE values (greener shades). A second observation is the reduced transferability of models trained on a location with extreme or highly distinct climate, such as Vancouver, Victoria, or Whitehorse. These locations belong to either very mild or very severe climate zones, making them less representative of the broader distribution of weather patterns across other zones. Consequently, models trained on these locations exhibit limited generalization to dissimilar climates.

From a building physics perspective, these observations can be attributed to differences in the dominant heat balance mechanisms governing energy demand across climate regimes. Locations within the same climate zone are exposed to comparable ranges of outdoor temperature, solar irradiance, wind speed, and humidity, leading to similar envelope heat transfer processes and internal load interactions. As a result, models trained within these zones learn weather-energy relationships that are physically consistent and easier transferable.

In contrast, models trained on climates characterized by thermally constrained operating regimes, such as subarctic zones dominated by prolonged heating demand (e.g., Whitehorse) or mild coastal zones with limited thermal extremes (e.g., Vancouver), exhibit limited generalization to other climates. In subarctic regions, building energy demand is primarily governed by steady conductive and convective heat losses driven by persistently low outdoor temperatures, reduced solar gains, and elevated wind-induced infiltration. This results in heating-dominated, low-variance energy profiles with limited excitation of cooling or transitional dynamics. Similarly, in mild coastal climates, moderate temperatures, high thermal inertia, and maritime effects suppress extreme heating and cooling loads, yielding smoothed demand profiles with weak sensitivity to short-term weather fluctuations. These constrain the model's ability to learn the full spectrum of weather-energy interactions, particularly nonlinear responses arising from seasonal transitions, solar-driven gains, and mixed-mode operation. Consequently, when applied to climates with stronger seasonal contrasts or competing thermal drivers, these models experience a regime mismatch: the learned representations have lower exposure to the coupled heating-cooling-weather dynamics that dominate energy use elsewhere. Such physical mismatches in dominant heat transfer mechanisms and excitation richness tend to reduce the transferability of learned temporal patterns across climate zones.

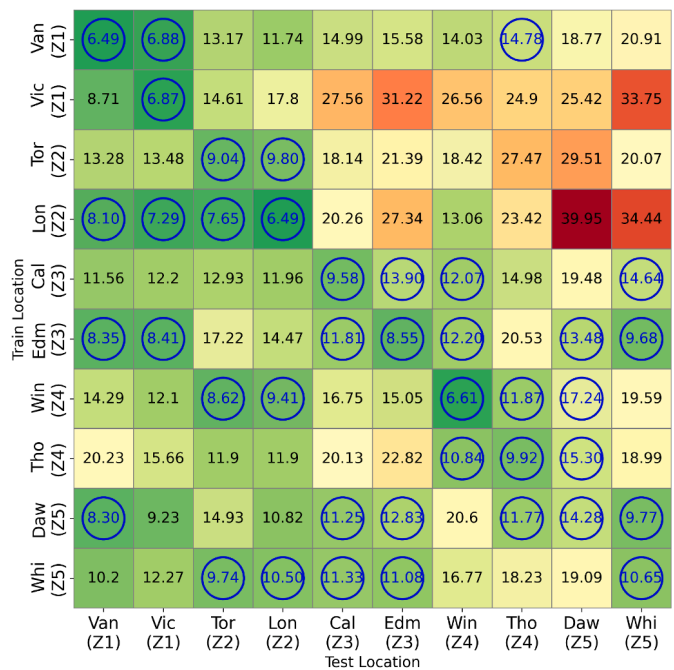

**Fig. 5.** TCN (Approach 1): SMAPE values (%) for cross-location model evaluation. Rows correspond to the training location and columns correspond to the test location. For each test location, the top 4 SMAPE values are circled. High intra-zone generalization performance observed.

Viewed from a machine learning perspective, this can also be understood more intuitively as a data imbalance problem. Training data from extreme climates are heavily skewed toward a single regime (e.g., constant heating), making it less representative of more diverse operating patterns. As a result, the model lacks sufficient exposure to varied inputs during training, reducing its ability to generalize across locations with broader and more dynamic climatic conditions.

Models trained on locations from Zones 2 to 4-which exhibit a wider range of seasonal variability and mixed weather conditions-tend to generalize more effectively across different climate zones. These locations expose the model to diverse patterns during training, enabling the development of more scalable and transferable representations. This is reflected in the heatmaps, where training on Zone 2 to Zone 4 locations yields consistently lower SMAPE values (greener regions) across various test locations, compared to models trained on Zone 1 (mild/coastal) or Zone 5 (extreme cold) locations. Another insight is that transferability generally degrades with increasing climatic dissimilarity between training and test zones. This is visually evident in the heatmaps, particularly for the Transformer model, where colors transition from green to red as one moves farther from the diagonal. This gradient highlights the

**Table 4**

SMAPE for the three surrogate modeling techniques across source-target location pairs with single location training. Format: Weekly (Annual). The best metrics for a train-test pair across the three modeling techniques are highlighted.

| Model | Zone | Train Loc | Test Loc | | | | | | | | | |
|---|---|---|---|---|---|---|---|---|---|---|---|---|
| | | | Zone 1 | | Zone 2 | | Zone 3 | | Zone 4 | | Zone 5 | |
| | | | Vancouver | Victoria | Toronto | London | Calgary | Edmonton | Winnipeg | Thompson | Dawson | Whitehorse |
| TCN | Z1 | Van | 6.49 (2.39) | 6.88 (2.22) | 13.17 (10.24) | 11.74 (3.42) | 14.99 (11.48) | 15.58 (4.05) | 14.03 (9.06) | 14.78 (3.43) | 18.77 (3.92) | 20.91 (15.54) |
| | | Vic | 8.71 (4.66) | 6.87 (2.59) | 14.61 (9.65) | 17.80 (18.51) | 27.56 (31.43) | 31.22 (40.65) | 26.56 (36.69) | 24.90 (33.55) | 25.42 (37.04) | 33.75 (48.78) |
| | Z2 | Tor | 13.28 (11.27) | 13.48 (11.58) | 9.04 (2.01) | 9.80 (1.81) | 18.14 (4.93) | 21.39 (2.99) | 18.42 (10.29) | 27.47 (28.98) | 29.51 (28.89) | 20.07 (3.55) |
| | | Lon | 8.10 (4.14) | 7.29 (4.61) | 7.65 (2.52) | 6.49 (1.96) | 20.26 (3.20) | 27.34 (8.90) | 13.06 (12.62) | 23.42 (20.02) | 39.95 (29.19) | 34.44 (15.31) |
| | Z3 | Cal | 11.56 (7.24) | 12.20 (7.60) | 12.93 (5.32) | 11.96 (5.80) | 9.58 (3.65) | 13.90 (4.14) | 12.07 (4.24) | 14.98 (8.54) | 19.48 (5.82) | 14.64 (2.74) |
| | | Edm | 8.35 (2.78) | 8.41 (2.75) | 17.22 (14.51) | 14.47 (10.92) | 11.81 (2.07) | 8.55 (1.89) | 12.20 (8.07) | 20.53 (19.84) | 13.48 (10.66) | 9.68 (3.73) |
| | Z4 | Win | 14.29 (12.27) | 12.10 (11.10) | 8.62 (3.27) | 9.41 (2.74) | 16.75 (4.61) | 15.05 (4.13) | 6.61 (2.04) | 11.87 (2.12) | 17.24 (1.84) | 19.59 (15.64) |
| | | Tho | 20.23 (20.90) | 15.66 (16.22) | 11.90 (11.29) | 11.90 (10.75) | 20.13 (7.81) | 22.82 (18.39) | 10.84 (5.32) | 9.92 (2.53) | 15.30 (8.18) | 18.99 (8.39) |
| | Z5 | Daw | 8.30 (2.12) | 9.23 (5.61) | 14.93 (9.71) | 10.82 (6.08) | 11.25 (6.33) | 12.83 (11.73) | 20.60 (20.95) | 11.77 (10.51) | 14.28 (4.06) | 9.77 (5.76) |
| | | Whi | 10.20 (7.78) | 12.27 (11.45) | 9.74 (2.63) | 10.50 (4.22) | 11.33 (2.26) | 11.08 (8.96) | 16.77 (13.28) | 18.23 (17.29) | 19.09 (16.52) | 10.65 (2.31) |
| Transformer | Z1 | Van | 5.28 (1.19) | 5.86 (2.04) | 8.53 (4.94) | 10.06 (5.20) | 15.85 (11.59) | 20.78 (17.57) | 16.72 (17.04) | 26.57 (31.14) | 32.14 (36.83) | 25.67 (20.69) |
| | | Vic | 6.68 (1.74) | 6.05 (1.98) | 8.89 (2.22) | 10.53 (2.66) | 13.91 (5.55) | 13.73 (2.19) | 13.54 (9.99) | 17.81 (21.06) | 21.77 (23.26) | 13.95 (6.30) |
| | Z2 | Tor | 7.18 (1.34) | 6.67 (1.52) | 9.77 (7.30) | 8.78 (7.08) | 11.08 (8.54) | 12.55 (9.98) | 8.15 (4.21) | 11.40 (2.00) | 15.58 (2.70) | 22.30 (17.10) |
| | | Lon | 6.24 (1.02) | 5.16 (1.70) | 6.83 (2.24) | 7.21 (3.37) | 7.44 (2.42) | 8.06 (1.97) | 8.53 (7.47) | 15.86 (13.97) | 20.09 (17.72) | 9.85 (4.00) |
| | Z3 | Cal | 10.76 (9.88) | 10.51 (10.04) | 6.92 (1.88) | 7.20 (2.03) | 6.94 (1.69) | 6.11 (1.66) | 7.30 (3.53) | 9.66 (7.48) | 9.50 (6.49) | 7.78 (1.56) |
| | | Edm | 6.72 (4.28) | 6.78 (4.30) | 11.30 (10.01) | 9.71 (8.43) | 5.93 (2.76) | 6.38 (2.96) | 9.55 (7.27) | 9.42 (7.73) | 9.08 (6.79) | 6.64 (2.10) |
| | Z4 | Win | 8.38 (4.19) | 6.79 (4.03) | 6.82 (1.91) | 6.59 (2.41) | 8.17 (2.47) | 9.69 (4.74) | 5.49 (2.55) | 9.82 (2.61) | 11.12 (1.91) | 15.39 (11.72) |
| | | Tho | 8.83 (4.87) | 7.52 (5.04) | 8.31 (2.19) | 8.04 (2.62) | 7.52 (2.71) | 6.48 (2.09) | 7.90 (4.01) | 5.67 (2.52) | 9.31 (6.65) | 6.58 (2.03) |
| | Z5 | Daw | 8.91 (6.16) | 6.92 (3.73) | 8.66 (4.67) | 9.20 (2.28) | 19.91 (19.96) | 17.83 (18.18) | 8.35 (1.69) | 9.91 (8.14) | 8.69 (3.73) | 16.50 (17.25) |
| | | Whi | 21.93 (19.59) | 22.48 (20.94) | 17.03 (6.80) | 17.02 (8.74) | 15.34 (7.12) | 20.49 (13.58) | 26.42 (19.01) | 25.88 (23.93) | 29.89 (30.45) | 22.93 (21.06) |
| Autoencoder | Z1 | Van | 5.77 (3.12) | 8.48 (5.40) | 14.56 (14.18) | 17.45 (17.20) | 16.04 (2.75) | 18.86 (20.65) | 21.20 (21.89) | 35.56 (41.91) | 32.68 (37.48) | 26.24 (20.21) |
| | | Vic | 9.84 (7.90) | 11.62 (10.33) | 18.25 (19.29) | 18.04 (18.51) | 14.29 (10.71) | 19.61 (20.57) | 21.99 (25.13) | 34.90 (42.00) | 35.22 (43.28) | 22.01 (22.68) |
| | Z2 | Tor | 10.14 (9.13) | 9.25 (7.99) | 7.44 (2.38) | 7.39 (3.08) | 13.27 (12.08) | 13.09 (12.53) | 6.77 (1.15) | 8.63 (7.86) | 15.09 (16.18) | 12.44 (10.07) |
| | | Lon | 14.53 (10.28) | 10.34 (2.47) | 6.67 (1.21) | 6.39 (2.18) | 12.28 (10.47) | 10.07 (6.47) | 9.16 (7.43) | 10.63 (9.61) | 15.87 (17.06) | 12.74 (10.36) |
| | Z3 | Cal | 14.92 (10.24) | 19.08 (16.80) | 12.20 (1.23) | 15.58 (4.57) | 8.76 (1.65) | 15.85 (14.91) | 10.88 (1.43) | 16.93 (2.94) | 19.42 (10.52) | 16.04 (10.00) |
| | | Edm | 19.11 (17.57) | 16.69 (16.35) | 14.99 (11.70) | 12.52 (9.97) | 16.05 (15.67) | 9.06 (4.56) | 11.62 (8.48) | 7.98 (1.56) | 9.09 (6.72) | 9.10 (4.75) |
| | Z4 | Win | 9.50 (4.09) | 6.77 (1.51) | 6.51 (2.85) | 7.13 (1.30) | 32.22 (33.75) | 21.06 (22.87) | 8.85 (8.37) | 11.76 (4.48) | 27.58 (32.51) | 29.55 (30.62) |
| | | Tho | 8.29 (2.44) | 8.02 (1.05) | 9.37 (1.98) | 7.92 (2.40) | 8.89 (2.70) | 9.39 (6.34) | 9.28 (7.50) | 8.49 (4.21) | 7.60 (1.89) | 11.65 (3.02) |
| | Z5 | Daw | 8.93 (5.41) | 9.11 (6.78) | 9.33 (1.92) | 8.62 (2.78) | 9.62 (2.35) | 8.72 (4.16) | 10.06 (2.16) | 9.10 (1.23) | 12.19 (1.80) | 12.38 (9.39) |
| | | Whi | 18.15 (6.30) | 10.86 (4.86) | 15.96 (11.70) | 15.70 (11.32) | 8.73 (4.26) | 9.17 (4.91) | 18.23 (16.68) | 11.92 (11.25) | 10.56 (9.02) | 13.90 (8.62) |

**Table 5**

RMSE for the three surrogate modeling techniques across source-target location pairs with single location training. The best metrics for a train-test pair across the three modeling techniques are highlighted.

| Model | Zone | Train Loc | Test Loc | | | | | | | | | |
|---|---|---|---|---|---|---|---|---|---|---|---|---|
| | | | Zone 1 | | Zone 2 | | Zone 3 | | Zone 4 | | Zone 5 | |
| | | | Vancouver | Victoria | Toronto | London | Calgary | Edmonton | Winnipeg | Thompson | Dawson | Whitehorse |
| TCN | Z1 | Van | 0.254 | 0.2678 | 1.3652 | 1.0738 | 1.8983 | 3.1314 | 2.1011 | 3.4908 | 5.4508 | 12.155 |
| | | Vic | 3.2652 | 0.2651 | 4.1829 | 10.1916 | 16.829 | 52.8976 | 55.8194 | 50.0268 | 69.9026 | 127.8582 |
| | Z2 | Tor | 0.7918 | 0.7957 | 0.6704 | 0.8074 | 2.2842 | 3.6934 | 4.1551 | 10.7311 | 13.2429 | 4.2368 |
| | | Lon | 0.3641 | 0.3028 | 0.4147 | 0.3192 | 2.7823 | 5.6167 | 2.4752 | 8.297 | 16.3479 | 8.4749 |
| | Z3 | Cal | 0.8855 | 0.7333 | 1.3228 | 1.1501 | 0.753 | 1.8397 | 1.3871 | 3.2722 | 5.5936 | 2.5451 |
| | | Edm | 0.4069 | 0.4084 | 2.2733 | 1.5688 | 1.3278 | 0.7001 | 1.5644 | 4.7974 | 2.8481 | 1.3415 |
| | Z4 | Win | 1.2255 | 0.862 | 0.538 | 0.902 | 1.9661 | 2.6095 | 0.5365 | 3.1122 | 5.8906 | 10.9117 |
| | | Tho | 3.1114 | 1.8358 | 1.3826 | 1.0298 | 2.8478 | 4.9763 | 0.9423 | 1.2542 | 2.9992 | 3.5125 |
| | Z5 | Daw | 0.3614 | 0.5644 | 1.6608 | 0.7657 | 1.6591 | 2.477 | 4.1267 | 1.7302 | 2.8757 | 0.9446 |
| | | Whi | 0.6377 | 0.8476 | 0.7798 | 0.8465 | 0.9902 | 0.9183 | 3.3891 | 4.5899 | 5.8451 | 0.9923 |
| Transformer | Z1 | Van | 0.1887 | 0.1952 | 0.6446 | 0.8434 | 2.7455 | 5.7595 | 5.0949 | 14.5475 | 21.0649 | 8.3833 |
| | | Vic | 0.2785 | 0.2086 | 0.5794 | 0.6642 | 1.3695 | 2.3044 | 2.5469 | 8.2575 | 11.1155 | 3.2656 |
| | Z2 | Tor | 0.2987 | 0.2582 | 0.9746 | 0.8747 | 0.7876 | 1.2588 | 0.6683 | 2.2067 | 3.4486 | 3.8826 |
| | | Lon | 0.2111 | 0.1494 | 0.4132 | 0.4397 | 0.4373 | 0.781 | 1.13 | 4.5898 | 6.6568 | 1.4346 |
| | Z3 | Cal | 0.7101 | 0.6231 | 0.4152 | 0.372 | 0.3532 | 0.4083 | 0.6908 | 2.2654 | 2.5205 | 0.745 |
| | | Edm | 0.2469 | 0.2657 | 0.9106 | 0.6225 | 0.2627 | 0.414 | 0.8342 | 1.1921 | 1.5157 | 0.5006 |
| | Z4 | Win | 0.3899 | 0.2654 | 0.3761 | 0.365 | 0.6407 | 1.3034 | 0.3198 | 1.6156 | 2.2772 | 5.217 |
| | | Tho | 0.452 | 0.3708 | 0.5044 | 0.4373 | 0.5061 | 0.3886 | 0.877 | 0.5298 | 1.3124 | 0.5008 |
| | Z5 | Daw | 0.4409 | 0.2681 | 0.5274 | 0.5987 | 4.568 | 3.7887 | 0.7335 | 1.3329 | 1.2404 | 3.2448 |
| | | Whi | 3.0384 | 3.2265 | 2.3043 | 2.3444 | 1.663 | 3.3822 | 5.987 | 7.8786 | 13.3338 | 5.4882 |
| Autoencoder | Z1 | Van | 0.2094 | 0.358 | 2.0463 | 2.4973 | 1.9423 | 4.5377 | 6.8164 | 18.7948 | 18.7017 | 7.7404 |
| | | Vic | 0.5655 | 0.6718 | 2.4489 | 2.4658 | 2.2874 | 4.8141 | 6.0761 | 18.4773 | 22.1955 | 7.1395 |
| | Z2 | Tor | 0.6182 | 0.3785 | 0.4336 | 0.4479 | 1.6824 | 1.654 | 0.6154 | 1.3544 | 3.5372 | 1.8248 |
| | | Lon | 0.9284 | 0.5651 | 0.3284 | 0.3601 | 0.981 | 0.972 | 1.1412 | 1.4886 | 3.7276 | 1.0642 |
| | Z3 | Cal | 1.5177 | 2.2178 | 1.0254 | 1.7205 | 0.8054 | 1.9784 | 1.0144 | 3.3422 | 3.4923 | 2.3585 |
| | | Edm | 2.8721 | 2.2396 | 2.5443 | 1.4539 | 3.753 | 0.7796 | 1.6513 | 0.767 | 1.4829 | 1.0762 |
| | Z4 | Win | 0.4538 | 0.2451 | 0.3306 | 0.429 | 4.9477 | 4.0098 | 1.1078 | 2.114 | 11.4082 | 6.3733 |
| | | Tho | 0.3278 | 0.3089 | 0.6395 | 0.4618 | 0.7024 | 0.825 | 1.4306 | 1.1653 | 0.6677 | 1.5013 |
| | Z5 | Daw | 0.4642 | 0.4644 | 0.5203 | 0.4936 | 0.6728 | 0.7047 | 0.9506 | 0.7082 | 1.6243 | 1.3307 |
| | | Whi | 1.9833 | 0.7611 | 1.7821 | 1.5237 | 0.6126 | 0.8403 | 3.1456 | 2.0745 | 2.2127 | 1.3851 |

model's declining performance as the climate difference between the training and test locations increases.

Locations in Zones 2 to 4 experience a broad spectrum of seasonal operating conditions, including heating-dominated winters, cooling-driven summers, and extended transitional periods. This variability activates multiple heat transfer pathways, including envelope conduction, solar gains, ventilation and infiltration losses, and internal load interactions, resulting in a wide and physically meaningful range of energy responses. From a machine learning standpoint, training data drawn from these zones is inherently more balanced, capturing diverse thermal regimes rather than a single dominant operating mode. As a result, models trained on such data are able to learn more generalizable weather-energy relationships and exhibit improved transferability when applied to other locations with overlapping but distinct climatic characteristics.

The Transformer-based model consistently exhibited greener heatmap regions compared to the other two approaches, highlighting its superior transferability across climate zones (see Fig. 6). This reflects the model's capacity to capture complex temporal patterns relevant for generalization. In contrast, the autoencoder-based approach showed comparatively lower transferability, which is expected given that its learned embeddings are not optimized for the prediction task ((see Fig. 7)). Nonetheless, its performance remains competitive, particularly in scenarios with limited computational resources, as the prediction model is lightweight and easily scalable.

To further evaluate how the model leverages weather patterns, we visualize the weekly predicted versus actual energy consumption for a randomly selected building design over a full year. For this, the Transformer model trained on Edmonton is selected due to its strong overall performance. Fig. 8 presents the weekly prediction results across all ten test locations, allowing for a visual assessment of the model's generalization capability. These plots reveal how well the model captures seasonal energy trends, even under unfamiliar weather conditions. In addition to SMAPE scores, the Pearson correlation coefficients (shown for each test location in Fig. 8) offer quantitative evidence of temporal alignment between predicted and actual values. Higher correlation values indicate that the model accurately captures underlying weather-driven consumption patterns. These results suggest that Transformer-based surrogate models, even when trained on a single location, can effectively support design optimization tasks under diverse climate scenarios.

### 4.4. Two-location training evaluation

In the single-location training experiments, model performance exhibited a modest decline when applied to climatically distant or extreme locations. To mitigate this and enhance generalization, we explored a two-location training setup. The hypothesis is that combining data from two climatically diverse locations during training increases the variety of weather patterns encountered, thereby improving the model's adaptability. To evaluate this, we extended the experiments by selecting Toronto as a primary training location and pairing it with London (from the same climate zone) as well as with four additional cities, each from a different climate zone. This setup enables an analysis of how increased climatic representation during training influences cross-location transferability.

For all three surrogate modeling approaches, hyperparameter tuning was performed independently for each training pair, following the same procedure used in the single-location training experiments. The selected hyperparameters for each training combination are summarized in Tables 6–8.

**Table 6**
Best hyperparameters for the TCN-based model under paired-location training. Each row corresponds to an instance trained on a specific pair of locations. Each trained instance is then used to evaluate its performance on other locations that are not part of the training pair, to assess transferability.

| Loc | CNN Filters | Embed Dim | FFNN Size | Dropout | Learning Rate |
|---|---|---|---|---|---|
| Tor + Van | 80 | 16 | 160 | 0.100 | 0.00011 |
| Tor + Lon | 96 | 112 | 256 | 0.115 | 0.00150 |
| Tor + Edm | 32 | 16 | 192 | 0.149 | 0.00064 |
| Tor + Win | 48 | 112 | 96 | 0.124 | 0.00461 |
| Tor + Daw | 112 | 16 | 160 | 0.197 | 0.00222 |

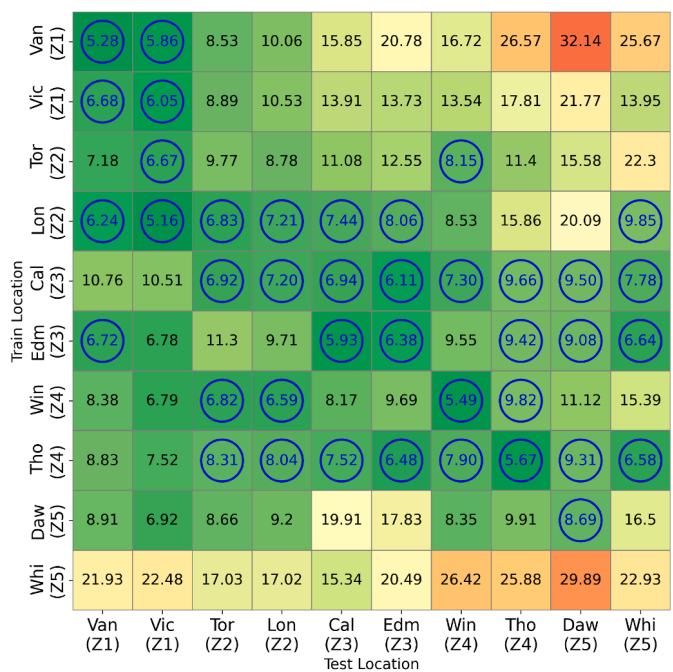

**Fig. 6.** Transformer (Approach 1): SMAPE values (%) for cross-location model evaluation. Rows correspond to the training location and columns correspond to the test location. Models trained on Zones 3 and 4 demonstrated superior generalization.

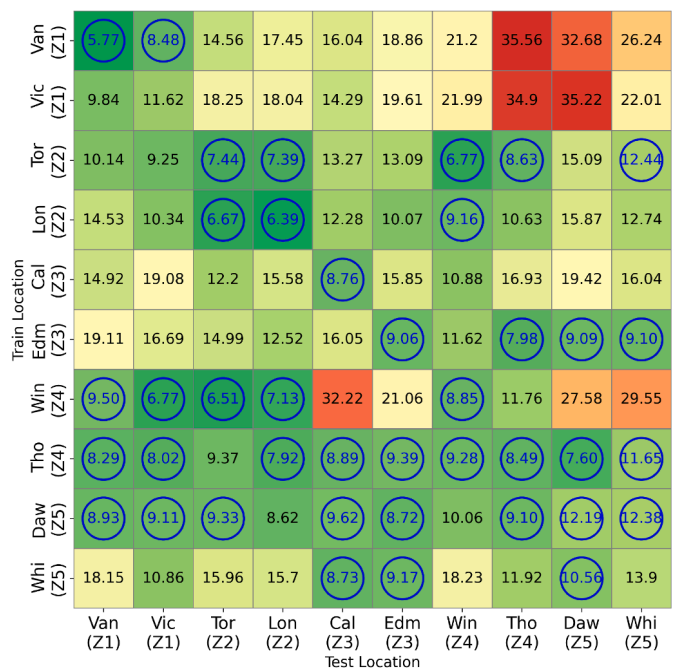

**Fig. 7.** Autoencoder (Approach 2): SMAPE values (%) for cross-location model evaluation. Rows correspond to the training location and columns correspond to the test location. SMAPE scores are generally higher for this approach compared to the other two, for the same train-test pairs.

**Table 7**
Best hyperparameters for the transformer-based model under paired-location training. Each row corresponds to an instance trained on a specific pair of locations. Each trained instance is then used to evaluate its performance on other locations that are not part of the training pair, to assess transferability.

| Loc | Embed Dim | Heads | T-FFNN Size | FFNN Size | Dropout | Learning Rate |
|---|---|---|---|---|---|---|
| Tor + Van | 8 | 1 | 57 | 227 | 0.120 | 0.00034 |
| Tor + Lon | 72 | 9 | 80 | 233 | 0.135 | 0.00071 |
| Tor + Edm | 96 | 12 | 116 | 233 | 0.129 | 0.00041 |
| Tor + Win | 8 | 1 | 58 | 174 | 0.200 | 0.00056 |
| Tor + Daw | 8 | 1 | 54 | 111 | 0.358 | 0.00255 |

**Table 8**
Best hyperparameters for the Autoencoder-based model under paired-location training. Each row corresponds to an instance trained on a specific pair of locations, Each trained instance is then used to evaluate its performance on other locations that are not part of the training pair, to assess transferability.

| Loc | Hidden Dim | Num Layers | Dropout | Learning Rate |
|---|---|---|---|---|
| Tor + Van | 492 | 4 | 0.382 | 0.00293 |
| Tor + Lon | 471 | 3 | 0.499 | 0.00821 |
| Tor + Edm | 450 | 4 | 0.498 | 0.00474 |
| Tor + Win | 285 | 3 | 0.367 | 0.00879 |
| Tor + Daw | 411 | 3 | 0.341 | 0.00987 |

Tables 9 and 10 summarize the SMAPE and MSE metrics obtained from the paired training setup. Compared to the single-location training results, the paired models exhibit improved generalizability and produce more stable predictions across different climate zones. This improvement is expected, as the inclusion of two geographically and climatically distinct training locations exposes the model to a wider range of weather conditions. Consequently, the model is better equipped to learn the relationship between weather variations and energy consumption, enhancing its ability to generalize to unseen environments.

The heatmaps in Fig. 9 further examine the benefits of paired-location training. Compared to single-location training, the green regions-indicating lower SMAPE values-are more extensive and consistent, particularly for the Transformer and TCN models. These stable regions extend beyond the diagonal axis, highlighting improved generalization to climate zones not seen during training. This effect is especially pronounced when the both paired training locations come from zones with a wide range of weather conditions, such as Z2, Z3, and Z4. The increased diversity in weather profiles appears to enhance the model's ability to generalize. This observation aligns with the earlier interpretation: training with two representative locations improves exposure to multiple operating regimes-heating, cooling, and transitional-and results in a more balanced distribution of weather-energy interactions in the dataset than using a single location. Together, these factors

**Table 9**

SMAPE for the three surrogate modeling techniques across source-target location pairs with two location training. Toronto was used as the primary training location, and each experiment paired it with one additional location to evaluate the impact of dual-location training. Format: Weekly (Annual). The best metrics for a train-test pair across the three modeling techniques are highlighted.

| Model | Zone | Paired Loc | Test Loc Zone 1 Vancouver | Victoria | Zone 2 Toronto | London | Zone 3 Calgary | Edmonton | Zone 4 Winnipeg | Thompson | Zone 5 Dawson | Whitehorse |
|---|---|---|---|---|---|---|---|---|---|---|---|---|
| TCN | Z1 | Van | 6.76 (3.22) | 7.22 (4.82) | 7.78 (1.24) | 6.76 (1.01) | 8.03 (1.92) | 8.94 (1.39) | 11.13 (8.99) | 15.06 (13.22) | 15.75 (13.65) | 11.02 (1.72) |
|  | Z2 | Lon | 6.32 (2.31) | 7.52 (3.24) | 6.86 (3.61) | 6.48 (1.52) | 14.20 (4.22) | 10.90 (2.62) | 8.53 (2.88) | 14.28 (9.46) | 16.92 (10.28) | 16.72 (2.13) |
|  | Z3 | Edm | 6.06 (1.61) | 5.83 (1.92) | 8.26 (6.69) | 7.12 (6.29) | 11.43 (10.56) | 9.58 (6.75) | 6.52 (2.50) | 9.65 (1.71) | 10.43 (1.75) | 9.92 (7.53) |
|  | Z4 | Win | 10.87 (7.89) | 12.00 (9.58) | 6.26 (2.39) | 5.49 (1.05) | 9.73 (4.89) | 10.26 (9.08) | 7.66 (4.17) | 9.88 (7.16) | 9.90 (5.15) | 14.86 (17.49) |
|  | Z5 | Daw | 17.47 (16.88) | 15.26 (14.18) | 7.46 (1.83) | 6.87 (1.82) | 17.62 (16.69) | 18.65 (13.91) | 12.29 (2.82) | 7.78 (2.19) | 16.56 (4.45) | 15.95 (10.80) |
| Trans. | Z1 | Van | 5.48 (2.40) | 5.03 (1.80) | 7.18 (1.37) | 8.07 (1.37) | 9.08 (3.04) | 10.32 (7.60) | 10.90 (10.37) | 22.78 (26.70) | 25.59 (30.00) | 17.24 (14.94) |
|  | Z2 | Lon | 6.10 (1.17) | 5.30 (1.48) | 5.53 (1.33) | 5.77 (1.87) | 12.55 (11.57) | 8.72 (6.81) | 6.26 (1.48) | 7.81 (3.96) | 8.21 (4.33) | 13.53 (11.09) |
|  | Z3 | Edm | 6.05 (3.65) | 6.16 (4.50) | 9.30 (6.95) | 9.97 (6.85) | 10.55 (9.93) | 9.74 (6.48) | 6.41 (3.47) | 6.63 (2.27) | 7.18 (1.73) | 7.38 (4.44) |
|  | Z4 | Win | 5.99 (2.14) | 5.46 (2.94) | 5.84 (2.73) | 5.69 (3.38) | 6.41 (2.88) | 7.73 (4.21) | 6.92 (6.08) | 10.02 (9.46) | 9.70 (4.19) | 12.49 (9.70) |
|  | Z5 | Whi | 10.08 (3.99) | 10.02 (5.55) | 11.60 (9.05) | 11.25 (9.73) | 17.94 (18.86) | 14.13 (14.88) | 9.30 (5.32) | 8.97 (1.79) | 11.08 (4.98) | 17.75 (18.41) |
| Auto. | Z1 | Van | 5.72 (2.11) | 6.14 (0.91) | 8.36 (2.94) | 6.99 (4.78) | 13.12 (7.48) | 13.16 (2.79) | 6.73 (3.62) | 7.30 (3.36) | 9.69 (3.98) | 15.10 (3.58) |
|  | Z2 | Lon | 6.98 (0.87) | 8.43 (4.41) | 7.35 (1.79) | 7.51 (2.56) | 14.11 (12.92) | 13.67 (3.90) | 7.91 (4.44) | 6.91 (1.78) | 10.20 (1.75) | 9.49 (3.51) |
|  | Z3 | Edm | 9.53 (8.71) | 8.33 (6.69) | 8.49 (4.56) | 9.03 (7.59) | 17.42 (16.59) | 14.98 (11.80) | 7.88 (2.63) | 14.04 (14.14) | 24.75 (29.35) | 10.91 (4.32) |
|  | Z4 | Win | 10.42 (8.92) | 9.52 (7.00) | 7.28 (2.25) | 8.30 (1.56) | 30.32 (31.48) | 25.25 (25.21) | 7.11 (4.32) | 15.10 (6.90) | 24.63 (23.71) | 26.09 (24.39) |
|  | Z5 | Daw | 13.22 (8.23) | 9.80 (6.46) | 7.92 (3.36) | 6.91 (3.64) | 8.85 (6.11) | 9.28 (3.05) | 6.27 (1.17) | 5.19 (1.74) | 9.54 (1.65) | 12.79 (10.53) |

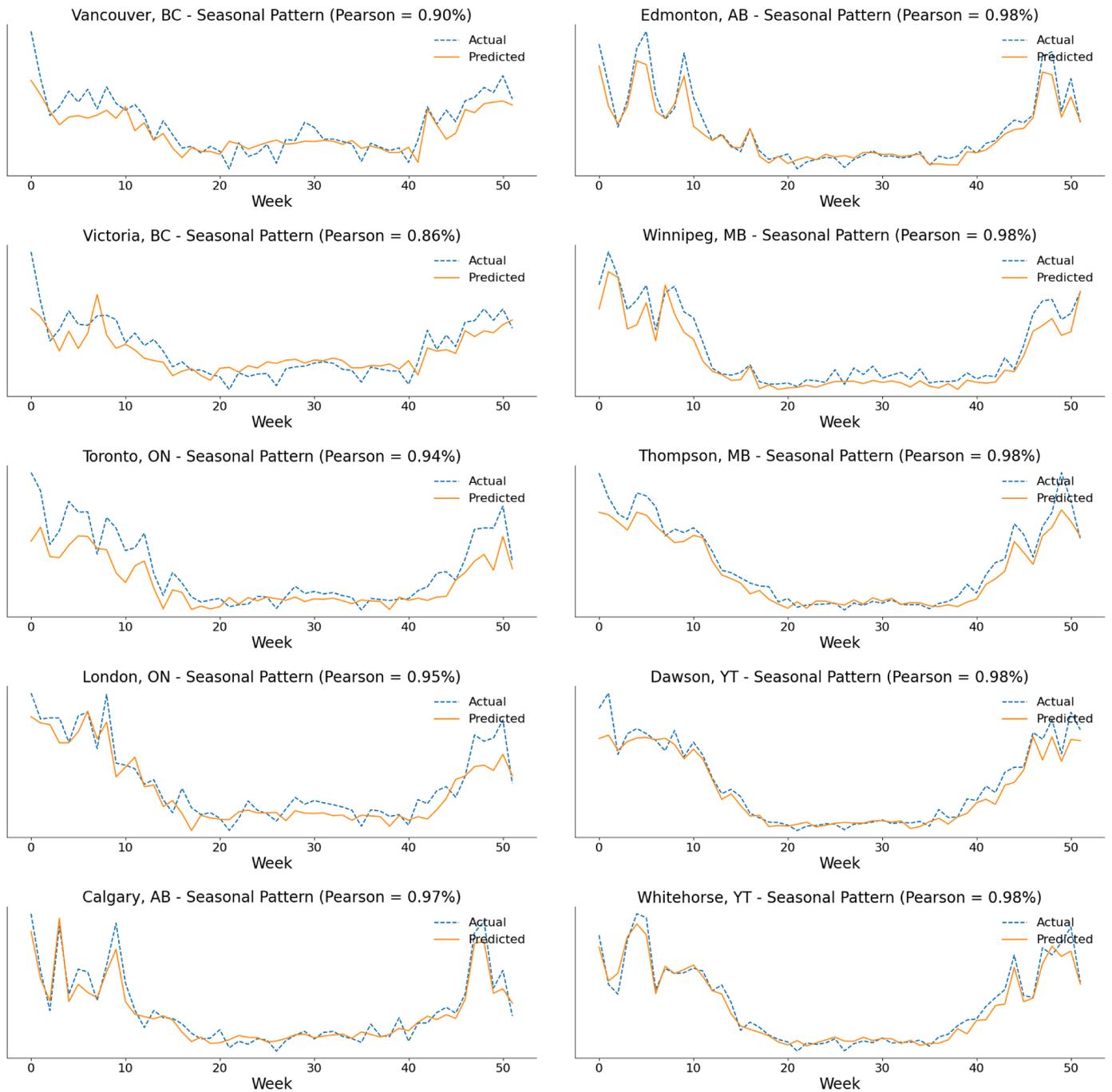

**Fig. 8.** Predicted vs. actual weekly energy consumption for the transformer model trained on Edmonton, evaluated across 10 test locations. Each plot compares the predicted and actual weekly patterns for one location.

strengthen the model's ability to learn generalized patterns that transfer across diverse climate zones.

When trained individually on London or Toronto, as seen in Figs. 5, and 6, the TCN and Transformer models exhibit noticeable performance degradation in Zones 3, 4, and 5, with accuracy declining as the climatic dissimilarity between the training and target zones increases. In contrast, paired training on both London and Toronto, as seen in Fig. 9, leads to more stable performance across all zones. This improvement is attributed to the increased training set size and the richer diversity of weather representations captured through the pairing. Similarly, combining Toronto with Vancouver enhances generalization across Zones 1 to 3, owing to the expanded representation of mild-weather patterns shared between the two locations. Moreover, pairing Toronto with Edmonton or Winnipeg yields consistently strong performance across all ten test locations, likely due to the balanced coverage of both moderate and extreme weather profiles provided by these cities.

These results demonstrate that combining data from two locations is beneficial; however, the effectiveness of this approach depends on the strategic selection of training locations. By carefully selecting complementary locations, the training set can be enriched with diverse weather patterns, thereby enhancing the model's ability to generalize to unseen target zones. While increasing the number of training locations further improves representational diversity and generalizability, it also incurs higher computational costs. Therefore, a well-designed training strategy that selects locations to maximize climatic diversity while minimizing redundancy offers an efficient trade-off between model performance and computational expense.

Table 10
RMSE for the three surrogate modeling techniques across source-target location pairs with two location training. The best metrics for a train-test pair across the three modeling techniques are highlighted.

| Model | Zone | Paired Loc | Test Loc | | | | | | | | | |
|---|---|---|---|---|---|---|---|---|---|---|---|---|
| | | | Zone 1 | | Zone 2 | | Zone 3 | | Zone 4 | | Zone 5 | |
| | | | Vancouver | Victoria | Toronto | London | Calgary | Edmonton | Winnipeg | Thompson | Dawson | Whitehorse |
| TCN | Z1 | Van | 0.2601 | 0.3508 | 0.4035 | 0.3724 | 0.5822 | 1.0904 | 2.0339 | 4.1975 | 5.1641 | 1.7473 |
| | Z2 | Lon | 0.2329 | 0.2965 | 0.3341 | 0.3494 | 1.4167 | 1.0085 | 1.03 | 3.6321 | 4.8489 | 2.2913 |
| | Z3 | Edm | 0.2482 | 0.225 | 0.533 | 0.6121 | 1.1121 | 0.7454 | 0.5065 | 1.6712 | 2.0377 | 1.0075 |
| | Z4 | Win | 0.5803 | 0.652 | 0.3133 | 0.2425 | 0.8285 | 1.8992 | 0.8319 | 1.906 | 2.4658 | 8.9998 |
| | Z5 | Daw | 1.1238 | 0.9052 | 0.4466 | 0.4149 | 2.9758 | 2.8575 | 1.8247 | 0.6629 | 3.8957 | 2.7066 |
| Trans. | Z1 | Van | 0.199 | 0.1437 | 0.3935 | 0.5471 | 0.7375 | 1.6492 | 2.3123 | 10.9719 | 14.9816 | 4.7433 |
| | Z2 | Lon | 0.2447 | 0.1658 | 0.2076 | 0.2536 | 1.3487 | 0.7888 | 0.4303 | 1.4964 | 1.9198 | 2.0511 |
| | Z3 | Edm | 0.2282 | 0.2203 | 1.3259 | 1.5363 | 1.6266 | 1.2042 | 0.7871 | 0.7887 | 1.0725 | 0.7832 |
| | Z4 | Win | 0.2083 | 0.1548 | 0.2318 | 0.2367 | 0.3759 | 0.7884 | 0.647 | 2.1789 | 2.0984 | 3.1231 |
| | Z5 | Daw | 0.5225 | 0.461 | 1.5971 | 1.5695 | 3.1898 | 2.2775 | 0.9648 | 0.7209 | 1.4728 | 3.3926 |
| Auto. | Z1 | Van | 0.2076 | 0.1865 | 0.5689 | 0.5808 | 1.0235 | 2.0032 | 0.483 | 0.6375 | 1.179 | 1.9713 |
| | Z2 | Lon | 0.2572 | 0.542 | 0.4847 | 0.5136 | 1.1265 | 2.4732 | 1.1044 | 0.714 | 1.2912 | 0.6596 |
| | Z3 | Edm | 0.4732 | 0.2999 | 0.5852 | 1.1699 | 2.3448 | 2.8395 | 0.8381 | 3.1659 | 11.1563 | 1.7855 |
| | Z4 | Win | 0.5251 | 0.3827 | 0.5692 | 0.6141 | 4.3358 | 3.6367 | 0.8538 | 2.5295 | 5.0373 | 4.2297 |
| | Z5 | Daw | 1.2066 | 0.6339 | 0.4957 | 0.4308 | 0.5567 | 1.4779 | 0.4899 | 0.3543 | 1.2511 | 1.2466 |

### 4.5. Evaluating the impact of strategic multi-zone representation

The previous section demonstrated that paired-location training improves generalizability by introducing more diverse weather patterns into the training data. However, the benefits of combining locations depend not only on the number of locations but also on which zones are included. To further investigate the role of climatic diversity in shaping a transferable surrogate model, we evaluate a *strategically mixed* training setup that combines three locations drawn from distinct climate zones. This strategy introduces the model to a wider variety of representative operating regimes from the selected locations. From data perspective, the inclusion of strategic climatically distinct locations promotes a more balanced training dataset, inducing greater transferability to similar locations.

Specifically, we select one location from each of the three zones: an extreme climate zone (Z5), a milder climate zone (Z1), and a middle-range zone (Z3). This combination aims to maximize the variability of weather conditions observed during training, enabling the model to better capture broad temporal weather-energy relationships. By systematically analyzing the impact of such strategically mixed representation, we assess whether a small but diverse subset of training locations can yield scalable surrogate models that generalize well across all climate zones-potentially reducing the need for exhaustive, full-region training while keeping computational costs manageable.

As shown in the heatmap in Fig. 10, with three well-selected training zones, the model achieves robust performance across all the zones, supporting the development of scalable and generalizable surrogate models. The Transformer model consistently achieves the lowest SMAPE across test locations, demonstrating superior generalization performance. This trend is also evident in Fig. 11, where the predicted values closely follow the seasonal variations observed in the actual data for a randomly selected design combination. The alignment between predicted and actual patterns is noticeably stronger compared to the single-location training results across all climate zones shown in Fig. 8. All models benefit from the systematic selection of climatically diverse training locations, enabling them to learn more transferable weather-to-energy relationships using only a minimal number of representative locations.

### 4.6. Comparison to previous studies

Previous studies have employed surrogate models for annual energy consumption prediction using full-year weather data as input features [14,17], making them a baseline for benchmarking our approach. To ensure a fair benchmark, we implemented a TCN-based surrogate model using the same architecture as our proposed TCN, but trained it to predict annual consumption directly using annual weather data. This allows us to isolate the effect of the prediction strategy itself, as the architecture remains unchanged. The other two approaches-Transformer and Autoencoder-are evaluated solely under the weekly prediction framework, enabling a consistent comparison across the three model types. These comparisons highlight the effectiveness of our proposed weekly prediction strategy over the conventional annual-level approach, and also demonstrate the potential of modeling techniques-such as Transformer and Autoencoder architectures-which, to our knowledge, have not been extensively explored in this context.

Fig. 12 presents heatmaps comparing the annual SMAPE scores of the annual-level TCN model, and our proposed weekly prediction approach. As evident from the heatmaps, models trained at the annual level struggle to generalize, particularly across climate zones, and in some cases, even within the same zone. This is reflected in the prevalence of red cells, indicating higher prediction error and poor transferability. They have to be trained with a handful of locations to learn underlying weather patterns.

In contrast, our proposed weekly prediction strategy demonstrates much better generalization, even when trained on data from a single location. In particular, the Transformer model trained on a Zone 3 or Zone 4 location exhibits stable and reliable performance across all climate zones. These results highlight the strong potential of our weekly prediction approach for developing scalable and transferable surrogate models capable of supporting diverse geographical settings with minimal training data.

## 5. Discussion

This section discusses the trade-offs of the proposed modeling approaches in terms of accuracy, generalization, and architectural design, followed by practical considerations related to scalability and deployment.

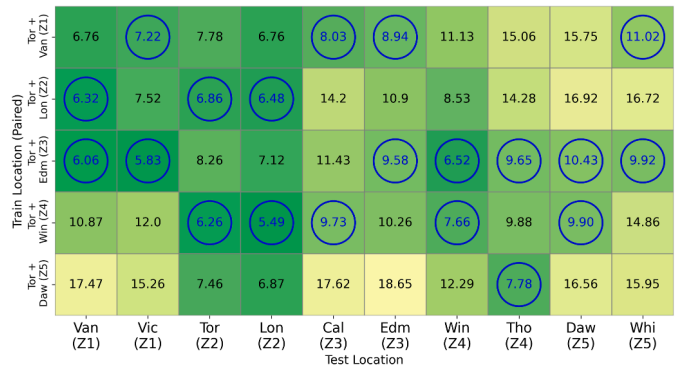

(a) Weekly SMAPE – TCN

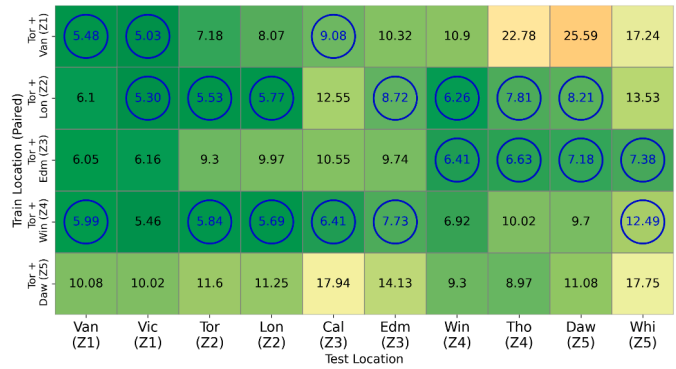

(b) Weekly SMAPE – Transformer

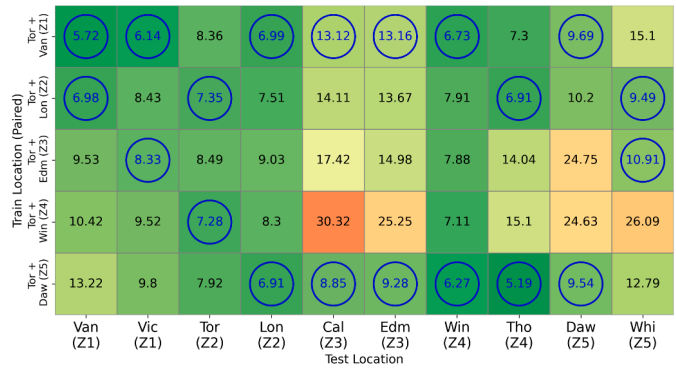

(c) Weekly SMAPE – Autoencoder

**Fig. 9.** SMAPE heatmaps for different surrogate modeling approaches. Each row corresponds to a model trained on the specified location pair, and each column represents the test location. For each test location, the top 2 models are circled.

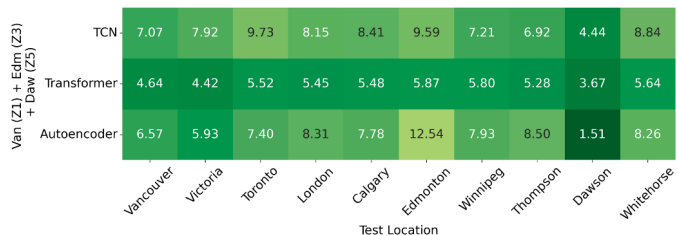

**Fig. 10.** SMAPE comparison for different surrogate modeling approaches. Each row corresponds to a model trained on a representative mix from Zones Z1, Z3, and Z5, with SMAPE scores reported for each test location (columns).

### 5.1. Trade-offs of the proposed techniques

The experimental results across all three modeling approaches-TCN, Transformer, and autoencoder-highlight important trade-offs between prediction accuracy, generalization capability, and architectural design.

*1) Joint Encoder Approaches (Approach 1).* The TCN and Transformer models, both trained in a supervised manner, demonstrated strong predictive performance when trained on regions with varied climatic profiles. Notably, models trained on Z3 or Z4 locations generalized well to most other locations, reinforcing the importance of including diverse weather patterns in training data. Between the two, the Transformer-based model consistently outperformed the TCN, particularly in capturing complex seasonal patterns and achieving lower SMAPE values across zones. Its attention mechanism enabled the model to better capture long-range temporal dependencies and nuanced weather effects.

*2) Autoencoder-Based Approach (Approach 2).* The autoencoder-based model introduces a two-stage pipeline where weather representations are learned independently through unsupervised training. Its ability to generalize to distant zones was more limited than the TCN or Transformer models. This limitation is partly due to the decoupling between the weather encoder and the energy prediction model, which may lead to embeddings that are not task specific. However, this approach's advantage of requiring less computationally intensive forward passes can be particularly beneficial in scenarios involving incremental learning and multi-location training, thereby partially offsetting its reduced accuracy.However, this approach's advantage of requiring less computationally intensive forward passes can be particularly beneficial in scenarios involving incremental learning and multi-location training, thereby partially offsetting its reduced accuracy.

*3) Impact of Multi-Location Training.* Across all approaches, models trained on two strategically selected locations showed notable improvements in generalization, especially when those locations offered contrasting but complementary weather characteristics. For instance, the Transformer model trained in Toronto and Edmonton consistently delivered strong performance across all test locations. This confirms that incorporating diverse weather signals in training helps the model learn more generalizable features. These results suggest that performance limitations observed with single-location models are not inherent to the modeling approach, but rather a result of limited training exposure. As more locations are incorporated into training-especially those covering a wider climatic spectrum-model performance can be expected to generalize more effectively. This highlights a promising path toward scalable surrogate modeling frameworks that can adapt to multiple climates with carefully selected representation during training.

In summary, the Transformer-based model offers the most balanced trade-off between accuracy and generalization. The autoencoder-based approach is particularly advantageous in scenarios involving incremental, multi-location training under limited computational resources. With increased training diversity, especially via multi-location representative dataset, all three methods show potential for improved robustness and wider applicability in surrogate-based building energy modeling.

### 5.2. Practical considerations and limitations

This subsection outlines key practical considerations and limitations that influence the deployment, scalability, and long-term applicability of the proposed transferable surrogate modeling methodology.

*1) Extensibility to Peak Load Prediction.* While this study focuses on predicting total weekly energy consumption, the proposed technique can be readily extended to include peak load prediction as an additional output. This extension requires no changes to the temporal encoder, weather representations, or overall architecture; only the simulation targets must be modified. For instance, peak weekly demand can be added alongside total energy consumption in a multi-output prediction setup. This would allow the surrogate model to support system sizing, demand-response planning, and other grid-impact evaluations, enhancing its utility in real-world building design and operational scenarios.

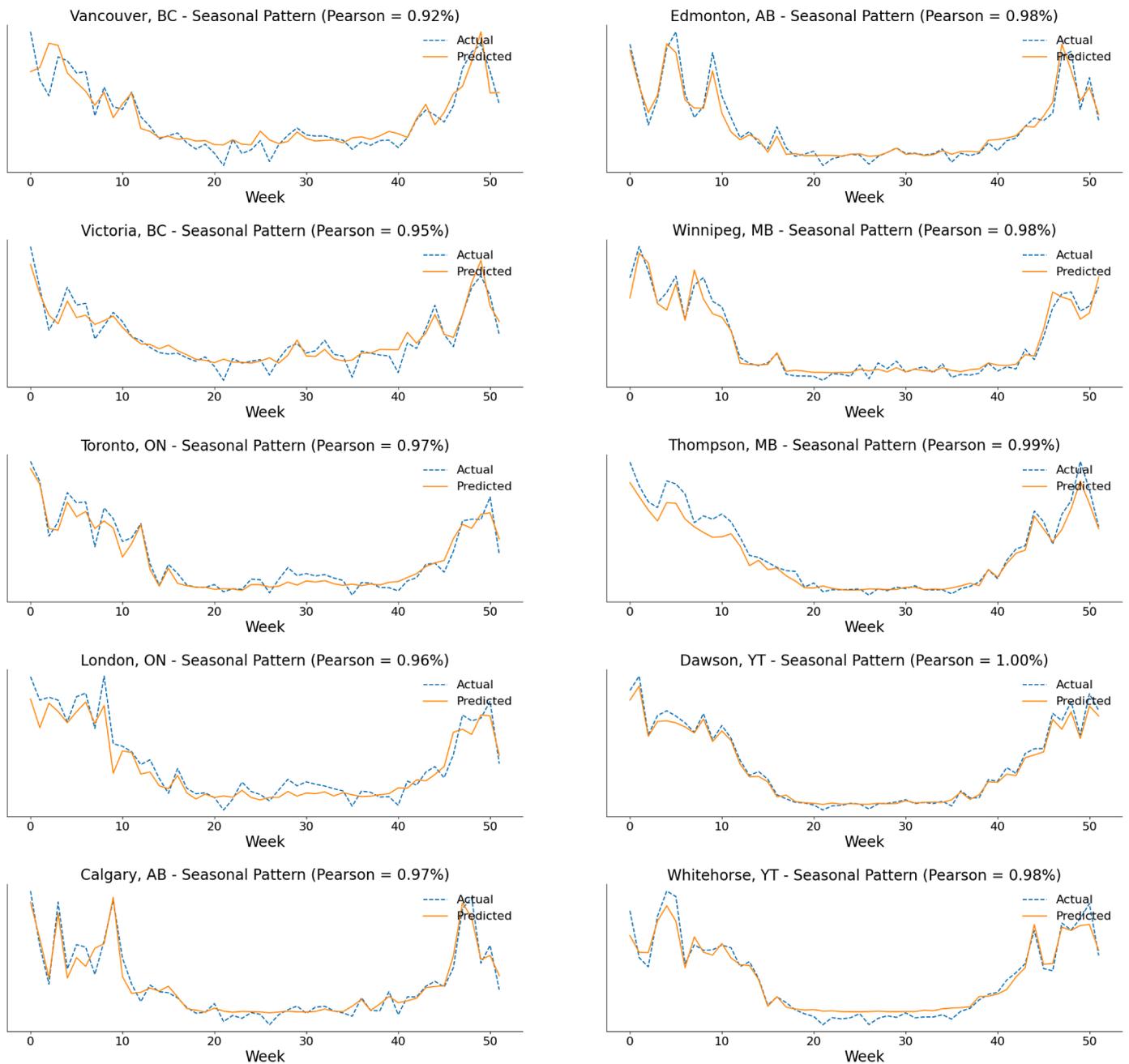

**Fig. 11.** Predicted vs. actual weekly energy consumption for the Transformer model trained on Vancouver + Edmonton + Dawson, evaluated across 10 test locations. Each plot compares the predicted and actual weekly patterns for one location.

*2) Data Availability.* While the proposed methodology enables scalable energy prediction across locations by training models on representative climates, its successful deployment still relies on the availability of high-quality weather input data. In particular, TMY weather files are required to represent the local climate conditions of the target location. Although TMY files are available for many regions through public repositories such as EnergyPlus and ASHRAE, their coverage is not universal. This limits the applicability of the methodology in regions lacking standardized or reliable weather data, particularly in remote, rural, or developing contexts.

*3) Computational Requirement.* The proposed transferable surrogate methodology reduces the simulation burden traditionally required for building design optimization. Once trained on a representative location, the surrogate can be directly used to evaluate design alternatives under different weather files, entirely bypassing the need for additional EnergyPlus simulations. This enables optimization to operate solely in the surrogate inference space, replacing thousands of simulation calls with near-instant predictions. As the model is only used during inference, it is computationally efficient and scalable, making it

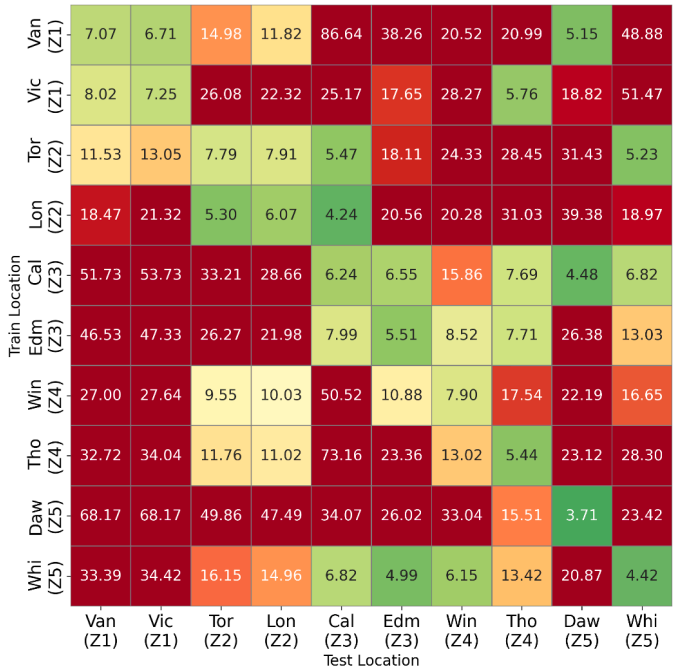
(a) Annual SMAPE – Annual Strategy

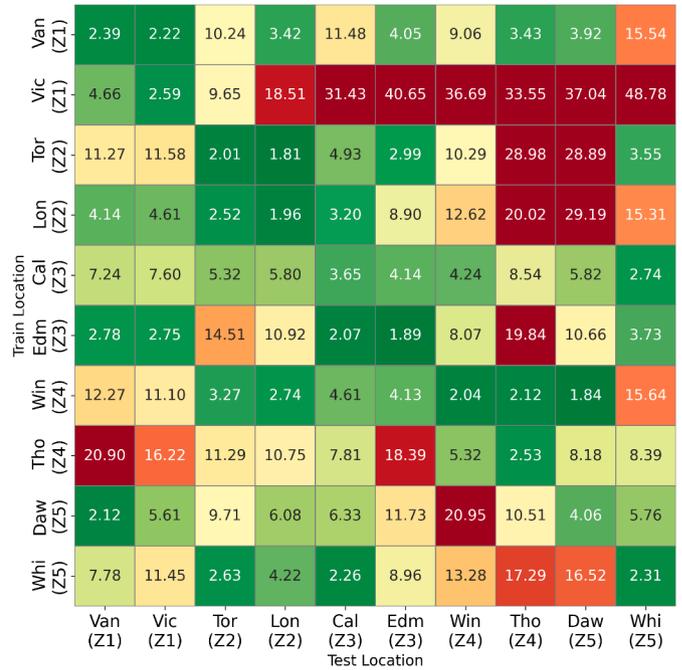
(b) Annual SMAPE – TCN Model

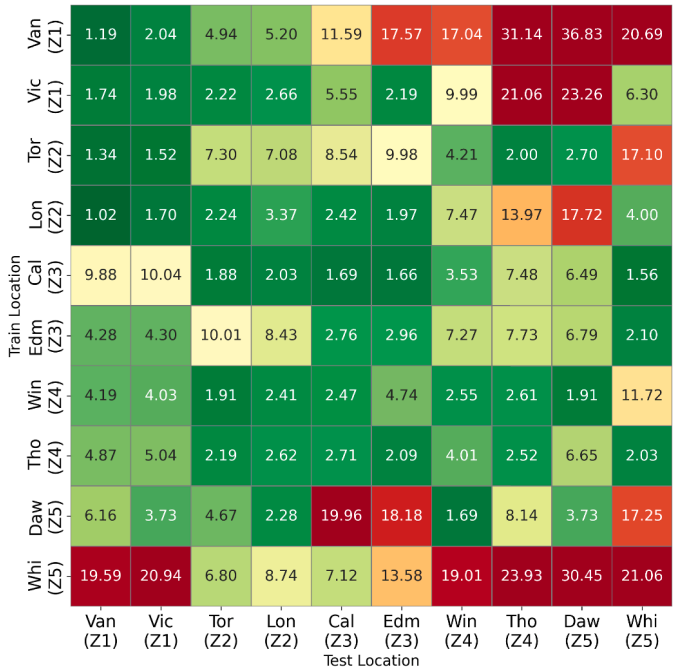
(c) Annual SMAPE – Transformer Model

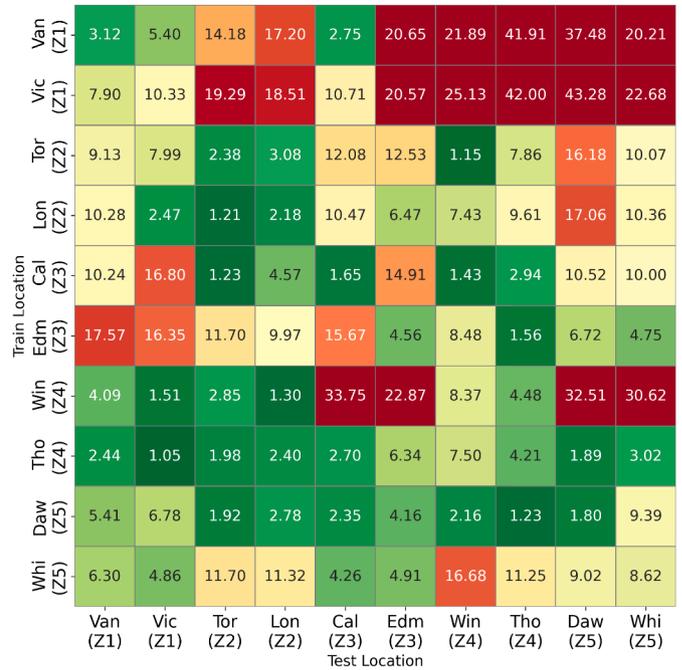
(d) Annual SMAPE – Autoencoder Model

**Fig. 12.** Annual and weekly SMAPE heatmaps for different surrogate modeling approaches across 10 locations. Lower values indicate better prediction accuracy and generalization.

particularly suited for time-sensitive or large-scale optimization workflows across diverse climates.

*4) Climate Change and Data Drift Considerations.* Future deployment scenarios must account for potential data drift arising from climate change. As buildings are exposed to increasingly frequent and intense extreme weather events, such as prolonged heatwaves or polar vortexes, weather patterns may deviate from those used in training. This can affect the validity of learned weather-energy relationships and reduce predictive accuracy. Future work could explore model adaptation strategies, to ensure continued reliability under evolving weather patterns.

## 6. Conclusion

This study proposed a high resolution weather-informed surrogate modeling approach aimed at reducing simulation requirements by leveraging commonly occuring short-term weather-energy demand patterns common to multiple locations. Through a comparative analysis of three approaches-TCN, Transformer, and an autoencoder-based model-we show how different proposed architectures handle the challenge of embedding weather patterns for this approach.

The findings demonstrate that, while performance varies by architecture and training region, all three models generalize intra- or interzones from training on just one or two well-chosen locations. This highlights the feasibility of building scalable surrogate models that minimize the simulation burden without sacrificing prediction accuracy using the commonly occurring high-resolution weather patterns across locations.

Looking ahead, this work opens the door to more flexible and reusable surrogate modeling strategies for building design optimization. Future directions include expanding location coverage, incorporating adaptive or online learning methods, and applying the approach to multi-objective optimization tasks and digital twin systems.

**Funding**

This work was supported by the Climate Action and Awareness Fund under Grant EDF-CA-2021i018, Environnement Canada, and by Alliance Grants (ALLRP) - Missions under Grant ALLRP 577133-2022. Computation was enabled in part by the Digital Research Alliance of Canada.

**Declaration of generative AI and AI-assisted technologies in the writing process**

During the preparation of this work, the authors used ChatGPT in order to improve the readability and language of the manuscript. After using this tool, the authors reviewed and edited the content as needed and take full responsibility for the content of the published article.

**CRediT authorship contribution statement**

**Piragash Manmatharasan:** Writing – original draft, Visualization, Software, Methodology, Investigation, Formal analysis, Data curation, Conceptualization; **Girma Bitsuamlak:** Writing – review & editing, Supervision, Funding acquisition, Conceptualization; **Katarina Grolinger:** Writing – review & editing, Supervision, Project administration, Methodology, Conceptualization.

**Data availability**

Data will be made available on request.

**Declaration of competing interest**

The authors declare the following financial interests/personal relationships which may be considered as potential competing interests:

Katarina Grolinger reports financial support was provided by Environment Canada. Katarina Grolinger reports equipment, drugs, or supplies was provided by Digital Research Alliance of Canada. If there are other authors, they declare that they have no known competing financial interests or personal relationships that could have appeared to influence the work reported in this paper.

**Supplementary material**

Supplementary material associated with this article can be found in the online version at 10.1016/j.enbuild.2026.117251.